\documentclass[acmsmall]{acmart}
\usepackage{times}
\usepackage{helvet}
\usepackage{courier}
\usepackage{graphicx}
\usepackage{algorithm}
\usepackage{algorithmic}
\usepackage{tabu}
\usepackage{array}
\usepackage{wrapfig}
\usepackage{dblfloatfix}
\usepackage{amsmath}
\renewcommand{\textrightarrow}{$\rightarrow$}
\usepackage[font=small,skip=0pt]{caption}
\def\BibTeX{{\rm B\kern-.05em{\sc i\kern-.025em b}\kern-.08emT\kern-.1667em\lower.7ex\hbox{E}\kern-.125emX}}

\begin{abstract}
\par\noindent\rule{\textwidth}{0.4pt}
With the widespread use of artificial intelligence (AI) systems and applications in our everyday lives,
accounting for fairness has gained significant importance in designing and engineering of such systems. AI systems can be used in many sensitive environments to make important and life-changing decisions; thus, it is crucial to ensure that these decisions do not reflect discriminatory behavior toward certain groups or populations. More recently some work has been developed in traditional machine learning and deep learning that address such challenges in different subdomains. With the commercialization of these systems, researchers are becoming more aware of the biases that these applications can contain and are attempting to address them. In this survey we investigated different real-world applications that have shown biases in various ways, and we listed different sources of biases that can affect AI applications. We then created a taxonomy for fairness definitions that machine learning researchers have defined in order to avoid the existing bias in AI systems. In addition to that, we examined  different domains and subdomains in AI showing what researchers have observed with regard to unfair outcomes in the state-of-the-art methods and ways they have tried to address them. There are still many future directions and solutions that can be taken to mitigate the problem of bias in AI systems. We are hoping that this survey will motivate  researchers to tackle these issues in the near future by observing existing work in their respective fields.  
\end{abstract}


\author{Ninareh Mehrabi}
\author{Fred Morstatter}
\author{Nripsuta Saxena}
\author{\\Kristina Lerman}
\author{Aram Galstyan, \small{USC-ISI}}
\renewcommand{\shortauthors}{Mehrabi et al.}

\begin{document}

\title{A Survey on Bias and Fairness in Machine Learning}

\maketitle
\par\noindent\rule{\textwidth}{0.4pt}

\section{Introduction}

Machine learning algorithms have penetrated every aspect of our lives. 
Algorithms make movie recommendations, suggest products to buy, and who to date. They are increasingly used in high-stakes scenarios such as loans \cite{mukerjee2002multi} and hiring decisions \cite{cohen2019efficient,bogen2018help}.
There are clear benefits to algorithmic decision-making; unlike people, machines do not become tired or bored~\cite{danziger2011extraneous,o2010routledge}, and can take into account orders of magnitude more factors than people can. However, like people, algorithms are vulnerable to biases that render their decisions ``unfair''~\cite{O'Neil:2016:WMD:3002861,angwin2016machine}.
In the context of decision-making, fairness is the \textit{absence of any prejudice or favoritism toward an individual or group based on their inherent or acquired characteristics}. Thus, an unfair algorithm is one 
whose decisions are skewed 
toward a particular group of people. A canonical example comes 
from a tool 
used by courts in the United States to make pretrial detention and release decisions. The software, Correctional Offender Management Profiling for Alternative Sanctions (COMPAS), measures the 
risk of a person to recommit another crime. 
Judges use COMPAS to decide whether to release an offender, or to keep him or her in prison. 
An investigation into the software found 
a bias against 
African-Americans:\footnote{https://www.propublica.org/article/machine-bias-risk-assessments-in-criminal-sentencing}
COMPAS is more likely to 
have higher false positive rates for African-American offenders than Caucasian offenders 
in falsely predicting them to be at a higher risk of recommitting a crime or recidivism.
Similar findings have been made in other areas, such as an AI system that judges beauty pageant winners but was biased against darker-skinned contestants,\footnote{https://www.theguardian.com/technology/2016/sep/08/artificial-intelligence-beauty-contest-doesnt-like-black-people} or facial recognition software in digital cameras that overpredicts Asians as blinking.\footnote{http://content.time.com/time/business/article/0,8599,1954643,00.html} These biased predictions stem from the hidden or neglected biases in data or algorithms.

In this survey we identify two potential sources of unfairness in machine learning outcomes---those that arise from biases in the data and those that arise from the algorithms. We review research investigating how biases in data skew what is learned by machine learning algorithms, and 
nuances in the way the algorithms themselves work to prevent them from making fair decisions---even when the data is unbiased. Furthermore, we observe that biased algorithmic outcomes might impact user experience, thus generating a feedback loop between data, algorithms and users that can perpetuate and even amplify existing sources of bias.

We begin the review with several highly visible real-world cases of where unfair machine learning algorithms have led to suboptimal and discriminatory outcomes in Section~\ref{sec:examples}. In Section~\ref{sec:loop}, we describe the different types and sources of biases that occur within the data-algorithms-users loop mentioned above. Next, in Section~\ref{sec:alg-fairness}, we present the different ways that the concept of fairness has been operationalized and studied in the literature. We discuss the ways in which these two concepts are coupled. Last, we will focus on different families of machine learning approaches, how fairness manifests differently in each one, and the current state-of-the-art for tackling them in Section~\ref{sec:methods}, followed by potential areas of future work in each of the domains in Section~\ref{sec:future}.

\section{Real-World Examples of Algorithmic Unfairness}
\label{sec:examples}
With the popularity of AI and machine learning over the past decades, and their prolific spread in different applications, safety and fairness constraints have become a significant issue for  researchers and engineers. Machine learning is used in courts to assess the probability that a defendant recommits a crime. It is used in different medical fields, in childhood welfare systems \cite{pmlr-v81-chouldechova18a}, and autonomous vehicles. All of these applications have a direct effect in our lives and can harm our society if not designed and engineered correctly, that is with considerations to fairness. \cite{osoba2017intelligence} has a list of the applications and the ways these AI systems affect our daily lives with their inherent biases, such as the  existence of bias in AI chatbots, employment matching, flight routing, and automated legal aid for immigration algorithms, and  search and advertising placement algorithms. \cite{howard2018ugly} discusses examples of how bias in the real world can creep into AI and robotic systems, such as bias in face recognition applications, voice recognition, and search engines. Therefore, it is important for researchers and engineers to be concerned about the downstream applications and their potential harmful effects when modeling an algorithm or a system. 
\subsection{Systems that Demonstrate Discrimination}
COMPAS is an exemplar of a discriminatory system. In addition to this, discriminatory behavior was also evident in an algorithm that would deliver advertisements promoting jobs in Science, Technology, Engineering, and Math (STEM) fields \cite{lambrecht2018algorithmic}. This advertisement was designed to deliver advertisements in a gender-neutral way. However, less women compared to men saw the advertisement due to gender-imbalance which would result in younger women being considered as a valuable subgroup and more expensive to show advertisements to. This optimization algorithm would deliver ads in a discriminatory way although its original and pure intention was to be gender-neutral. Bias in facial recognition systems \cite{raji2019actionable} and recommender systems \cite{schnabel2016recommendations} have also been largely studied and evaluated and in many cases shown to be discriminative towards certain populations and subgroups. In order to be able to address the bias issue in these applications, it is important for us to know where these biases are coming from and what we can do to prevent them. 

We have enumerated the bias in COMPAS, which is a widely used commercial risk assessment software. In addition to its bias, it also contains performance issues when compared to humans. When compared to non-expert human judgment in a study, it was discovered to be not any better than a normal human \cite{Dresseleaao5580}. It is also interesting to note that although COMPAS uses 137 features, only 7 of those were presented to the people in the study. \cite{Dresseleaao5580} further argues that COMPAS is not any better than a simple logistic regression model when making decisions. We should think responsibly, and recognize that the application of these tools, and their subsequent decisions affect peoples' lives; therefore, considering fairness constraints is a crucial task while designing and engineering these types of sensitive tools. In another similar study, while investigating sources of group unfairness (unfairness across different groups is defined later), the authors in \cite{tolan2019machine} compared SAVRY, a tool used in risk assessment frameworks that includes human intervention in its process, with automatic machine learning methods in order to see which one is more accurate and more fair. Conducting these types of studies should be done more frequently, but prior to releasing the tools in order to avoid doing harm.

\subsection{Assessment Tools}
An interesting direction that researchers have taken is introducing tools that can assess the amount of fairness in a tool or system. For example, Aequitas~\cite{saleiro2018aequitas} is a toolkit that lets users to test models with regards to several bias and fairness metrics for different population subgroups. Aequitas produces reports from the obtained data that helps data scientists, machine learning researchers, and policymakers  to make conscious decisions and avoid harm and damage toward certain populations. AI Fairness 360 (AIF360) is another toolkit developed by IBM in order to help moving fairness research algorithms into an industrial setting and to create a benchmark for fairness algorithms to get evaluated and an environment for fairness researchers to share their ideas  \cite{bellamy2018ai}. These types of toolkits can be helpful for learners, researchers, and people working in the industry to move towards developing fair machine learning application away from discriminatory behavior.

\section{Bias in Data, Algorithms, and User Experiences} 
\label{sec:loop}

\begin{figure}[!bt]
\includegraphics[width=\textwidth,trim=0cm 0cm 0cm 0cm,clip=true]{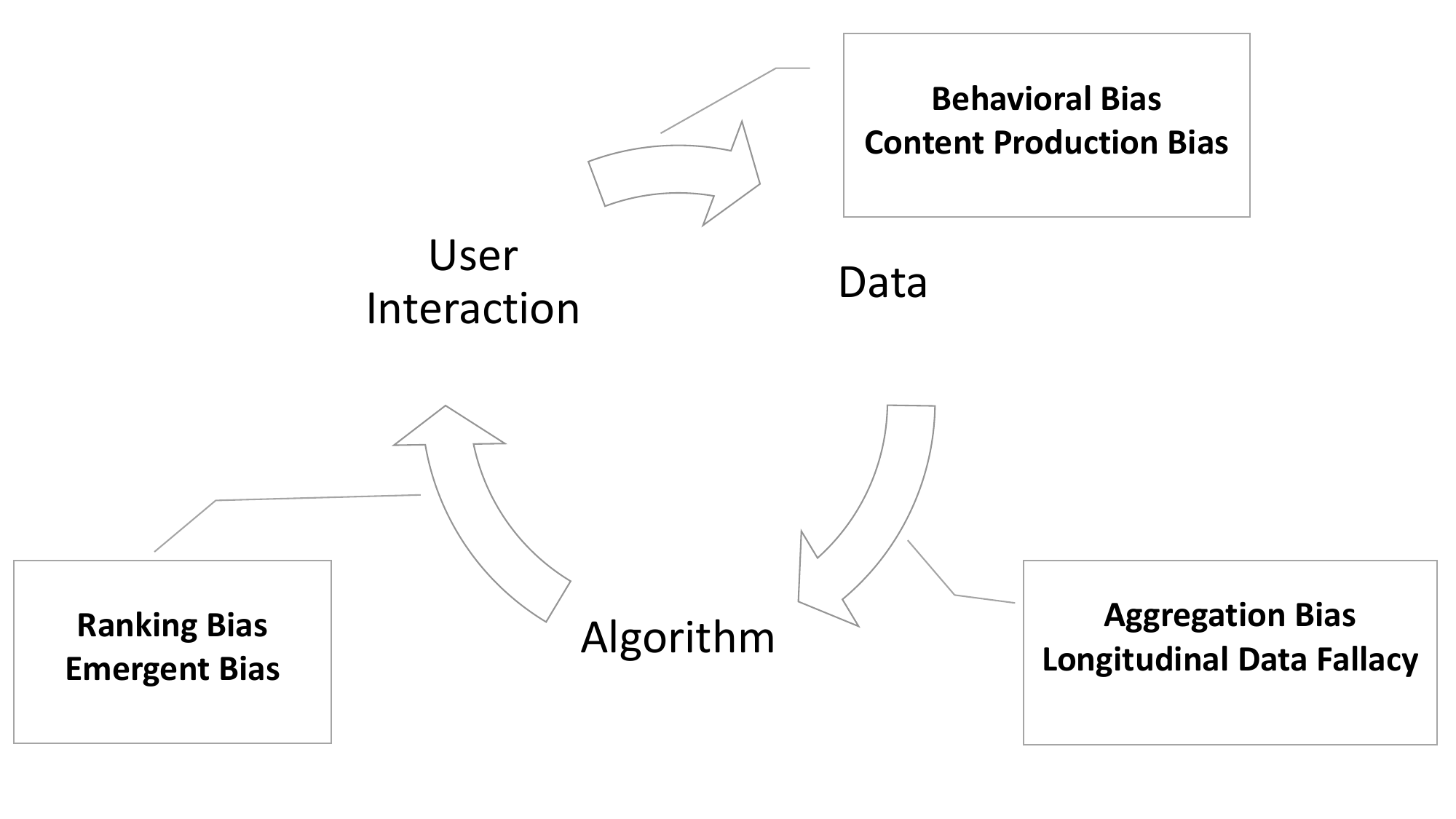}
\caption{Examples of bias definitions placed in the data, algorithm, and user interaction feedback loop.}
\label{fig:cycle}
\end{figure}

Most AI systems and algorithms are data driven and require data upon which to be trained. Thus, data is tightly coupled to the functionality of these algorithms and systems. In the cases where the underlying training data contains biases, the algorithms trained on them will learn these biases and reflect them into their predictions. As a result, existing biases in data can affect the algorithms using the data, producing biased outcomes. Algorithms can even amplify and perpetuate existing biases in the data. In addition, algorithms themselves can display biased behavior due to certain design choices, even if the data itself is not biased. The outcomes of these biased algorithms can then be fed into real-world systems and affect users' decisions, which will result in more biased data for training future algorithms. For example, imagine a web search engine 
that puts specific results at the top of its list. Users tend to interact most with the top results and pay little attention to those further down the list~\cite{Lerman14plosone}. The interactions of users with items will then be collected by the web search engine, and the data will be 
used to make future decisions on how information should be presented based on popularity and user interest. As a result, results at the top will become more and more popular, not because of the nature of the result but due to the biased interaction and placement of results by these algorithms~\cite{Lerman14plosone}. 
The loop capturing this feedback between biases in data, algorithms, and user interaction is illustrated in Figure~\ref{fig:cycle}. We use this loop to categorize definitions of bias in the section below.
\subsection{Types of Bias}
Bias can exist in many shapes and forms, some of which can lead to unfairness 
in different downstream learning tasks.
In \cite{suresh2019framework}, authors talk about sources of bias in machine learning with their categorizations and descriptions in order to motivate future solutions to each of the sources of bias introduced in the paper. In \cite{olteanu2016social}, the authors prepare a complete list of different types of biases with their corresponding definitions that exist in different cycles from data origins to its collection and its processing. Here we will reiterate the most important sources of bias introduced in these two papers and also add in some work from other existing research papers. Additionally, we will introduce a different categorization of these definitions in the paper according to the data, algorithm, and user interaction loop.
\subsubsection{Data to Algorithm}
In this section we talk about biases in data, which, when used by ML training algorithms, might result in biased algorithmic outcomes.
\begin{enumerate}
    \item \textbf{Measurement Bias.} \textit{Measurement, or reporting, bias arises from how we choose, utilize, and measure particular features} \cite{suresh2019framework}. An example of  this type of bias was observed in the recidivism risk prediction tool COMPAS, where 
    prior arrests and friend/family arrests were used as proxy variables to measure level of ``riskiness'' or ``crime''----which on its own can be viewed as mismeasured proxies. This is partly due to the fact that minority communities are controlled and policed more frequently, so they have higher arrest rates. However, one should not conclude that because people coming from minority groups have higher arrest rates therefore they are more dangerous as there is a difference in how these groups are assessed and controlled \cite{suresh2019framework}.

    \item \textbf{Omitted Variable Bias.} \textit{Omitted variable bias\textsuperscript{\ref{fl}} occurs when one or more important variables are left out of the model} \cite{clarke2005phantom,mustard2003reexamining,riegg2008causal}. An example for this case would be when someone designs a model to predict, with relatively high accuracy, the annual percentage 
    rate at which customers will stop subscribing to a service, but soon observes that the majority of users are canceling their subscription without receiving any warning from the designed model. Now imagine that the reason for canceling the subscriptions is appearance of a new strong competitor in the market which offers the same solution, but for half the price. The appearance of the competitor was something that the model was not ready for; therefore, it is considered to be an omitted variable. 

    \item \textbf{Representation Bias.} \textit{Representation bias arises from how we sample from a population during data collection process}  \cite{suresh2019framework}. Non-representative samples lack the diversity of the population, with missing subgroups and other anomalies. Lack of geographical diversity in datasets like ImageNet (as shown in Figures \ref{imagenet1} and \ref{imagenet2}) 
    results in demonstrable bias towards Western cultures.
    
     \item \textbf{Aggregation Bias.} \textit{Aggregation bias (or ecological fallacy) arises when false conclusions are drawn 
     about individuals from observing the entire population.} 
     An example of this type of bias can be seen in clinical aid tools. Consider diabetes patients who have apparent morbidity differences across ethnicities and genders. Specifically, HbA1c levels, that are widely used to diagnose and monitor diabetes, differ in complex ways across genders and ethnicities. Therefore,  
     a model that ignores individual differences will likely not be well-suited for all ethnic and gender groups in the population~\cite{suresh2019framework}.
     This is true even when they are represented equally in the training data. Any general assumptions about  
     subgroups within the population can result in aggregation bias.

    \begin{figure}[H]
    \centering
    \includegraphics[width=0.48\columnwidth]{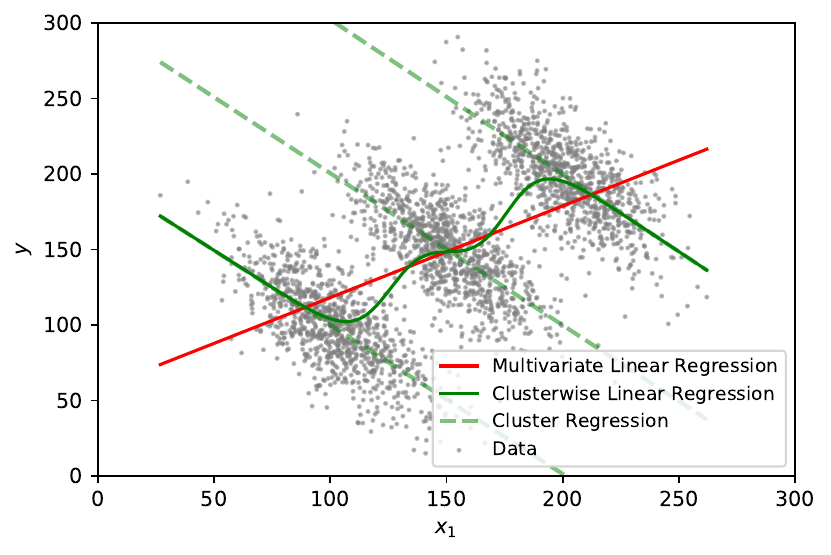}
    \includegraphics[width=0.48\columnwidth]{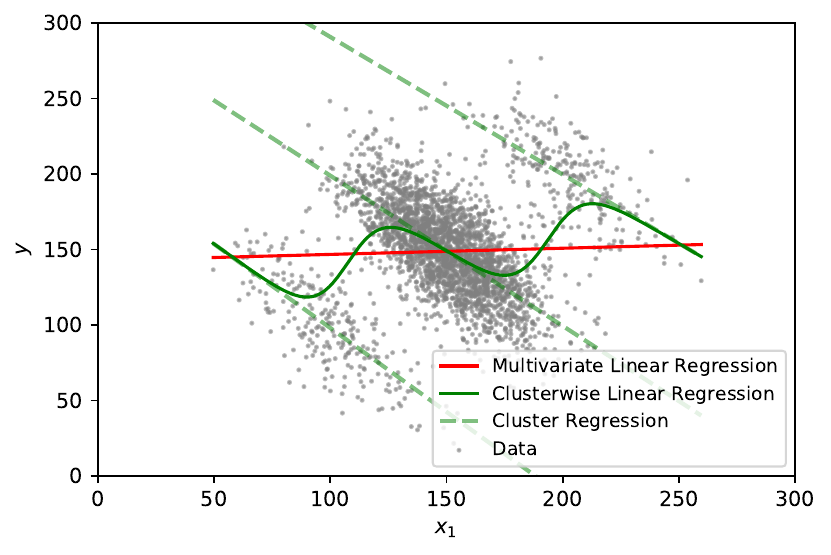}
    \caption{Illustration of biases in data.
    The red line shows the regression (MLR) for the entire population, while dashed green lines are regressions for each subgroup, and the solid green line is the unbiased regression. (a) When all subgroups are of equal size, then MLR shows a positive relationship between the outcome and the independent variable. (b) Regression shows almost no relationship  in less balanced data. The relationships between  variables within each subgroup, however, remain the same. (Credit: Nazanin Alipourfard) \label{fig:explain}}
    \end{figure}
\begin{figure}[h]
\includegraphics[width=0.75\textwidth, trim=2cm 5cm 0cm 6cm,clip=true]{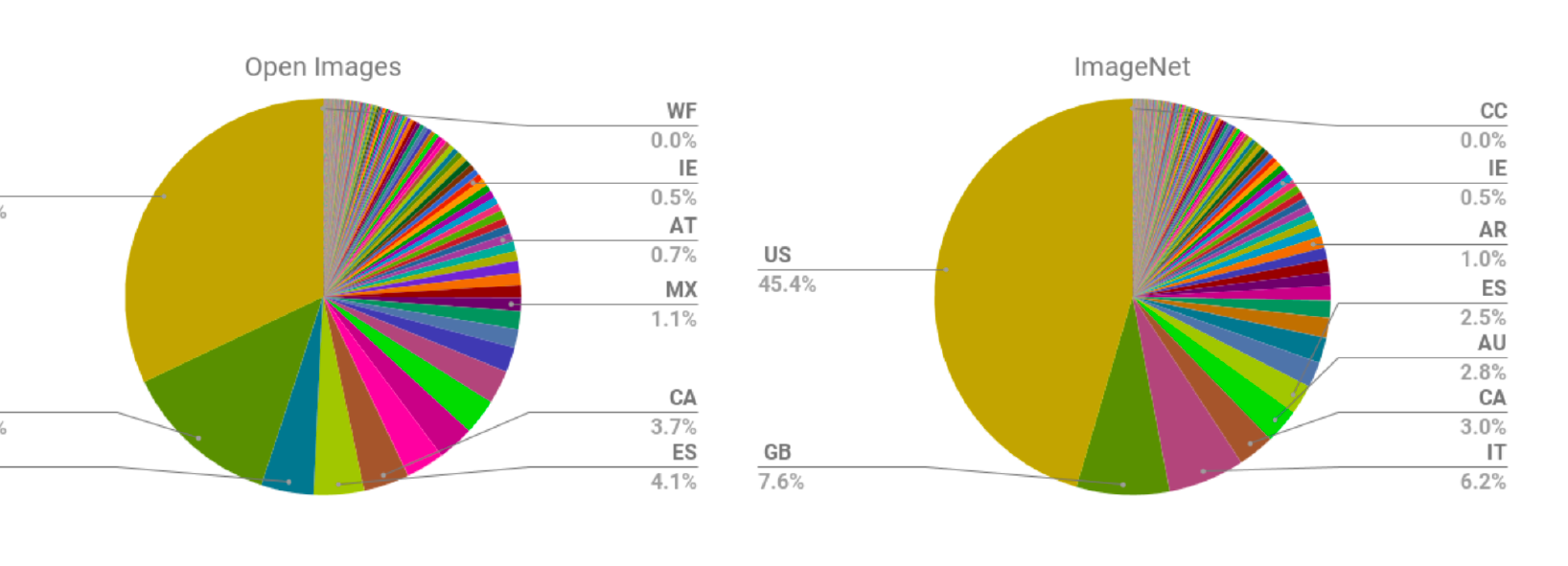}
\caption{Fraction of each country, represented by their two-letter ISO codes, in Open Images and ImageNet image datasets. In both datasets, US and Great Britain represent the top locations, from \cite{shankar2017no} \textsuperscript{\textcopyright} Shreya Shankar.}
\label{imagenet1}
\end{figure}
\begin{figure}[h]
\includegraphics[width=0.75\textwidth,trim=0cm 2cm 0cm 3.3cm,clip=true]{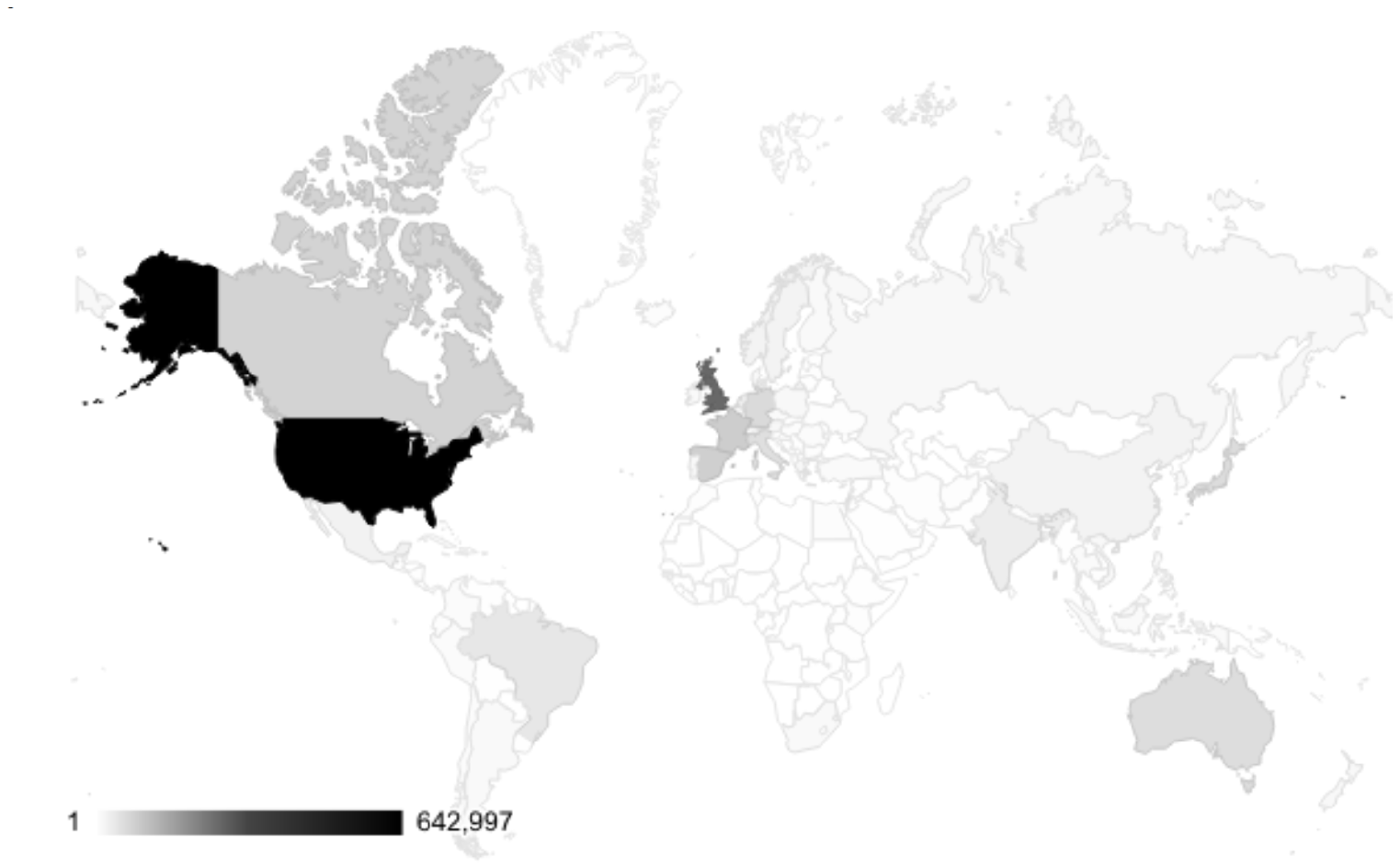}
\caption{Geographic distribution of countries in the Open Images data set. In their sample, almost one third of the data was US-based, and 60\% of the data was from the six most represented countries across North America and Europe, from \cite{shankar2017no} \textsuperscript{\textcopyright} Shreya Shankar.}
\label{imagenet2}
\end{figure}
 \begin{enumerate}
    \item \textbf{Simpson's Paradox.} Simpson's paradox is a type of aggregation bias that arises in the analysis of heterogeneous data~\cite{blyth1972simpson}. 
    The paradox arises when an association observed in aggregated data disappears or reverses when the same data is disaggregated into its underlying subgroups (Fig.~\ref{fig:explain}(a)). One of the better-known examples of the type of paradox arose during the gender bias lawsuit in university admissions against UC Berkeley~\cite{bickel1975sex}.
    After analyzing graduate school admissions data, it seemed like there was bias toward women, a smaller fraction of whom were being admitted to graduate programs compared to their male counterparts. However, when admissions data was separated and analyzed over the departments, women applicants had equality and in some cases even a small advantage over men. The paradox happened as women tended to apply to departments with lower admission rates for both genders. Simpson's paradox has been observed in a variety of domains, including biology~\cite{chuang2009simpson}, psychology~\cite{kievit2013simpson},  astronomy~\cite{minchev2019yule},  and computational social science~\cite{Lerman2018}.

    \item{\textbf{Modifiable Areal Unit Problem}} is a statistical bias in geospatial analysis, which arises when modeling data at different levels of spatial aggregation~\cite{Gehlke1934}. This bias results in different trends learned when data is aggregated at different spatial scales.
\end{enumerate}
   \item \textbf{Sampling Bias.} \textit{Sampling bias is similar to representation bias, and it arises due to {non-random} sampling of subgroups.} As a consequence of sampling bias, the trends estimated for one population may not generalize to data collected from a new population. For the intuition, consider the example in {Figure~\ref{fig:explain}}. The left plot represents data collected during a study from three subgroups, which were uniformly sampled (Fig.~\ref{fig:explain}(a)). Suppose the next time the  study was conducted, one of the subgroups was sampled more frequently than the rest (Fig.~\ref{fig:explain}(b)). The positive trend found by the regression model in the first study almost completely disappears (solid red line in plot on the right), although the subgroup trends (dashed green lines) are unaffected.
 
    \item \textbf{Longitudinal Data Fallacy.}
    Researchers analyzing temporal data must use \emph{longitudinal analysis} to track cohorts over time to learn their behavior. Instead, temporal data is often modeled using cross-sectional analysis, which combines diverse cohorts at a single time point. The heterogeneous cohorts can bias cross-sectional analysis, leading to different conclusions than longitudinal analysis. 
    As an example, analysis of bulk Reddit data~\cite{Barbosa2016} revealed that comment length decreased over time on average. However, bulk data represented a cross-sectional snapshot of the population, which in reality contained different cohorts who joined Reddit in different years. When data was disaggregated by cohorts, the comment length within each cohort was found to increase over time.

    \item \textbf{Linking Bias.}  \textit{Linking bias arises when network attributes obtained from user connections, activities, or interactions differ and misrepresent the true behavior of the users} \cite{olteanu2016social}. In \cite{mehrabi2019debiasing} authors show how social networks can be biased toward low-degree nodes when only considering the links in the network and not considering the content and behavior of users in the network. \cite{wilson2009user} also shows that user interactions are significantly different from social link patterns that are based on features, such as method of interaction or time. The differences and biases in the networks can be a result of many factors, such as network sampling, as shown in \cite{gonzalez2014assessing,morstatter2013sample}, which can change the network measures and cause different types of problems.

    \end{enumerate}
\subsubsection{Algorithm to User}
Algorithms modulate user behavior. Any biases in algorithms  might introduce biases in user behavior. In this section we talk about biases that are as a result of algorithmic outcomes and affect user behavior as a consequence.
\begin{enumerate}
 
 \item \textbf{Algorithmic Bias.} \textit{Algorithmic bias is when the bias is not present in the input data and is added purely by the algorithm}  \cite{Baeza-Yates:2018:BW:3229066.3209581}. The algorithmic design choices, such as use of certain optimization functions, regularizations, choices in applying regression models on the data as a whole or considering subgroups, and the general use of statistically biased estimators in algorithms \cite{danks2017algorithmic}, can all contribute to biased algorithmic decisions that can bias the outcome of the algorithms.
\item \textbf{User Interaction Bias.} \textit{User Interaction bias is 
    a type of bias that can not only be observant on the Web but also get triggered from two sources---the user interface and through the user itself by imposing his/her self-selected biased behavior and interaction }\cite{Baeza-Yates:2018:BW:3229066.3209581}. This type of bias can be influenced by other types and subtypes, such as presentation and ranking biases.
   
    \begin{enumerate}
    \item  \textbf{Presentation Bias.} \textit{Presentation bias is a result of how information is presented \cite{Baeza-Yates:2018:BW:3229066.3209581}.} For example, on the Web 
    users can only click on content that they see, so the seen content gets clicks, while everything else gets no click. And it could be the case that the user does not see all the information on the Web \cite{Baeza-Yates:2018:BW:3229066.3209581}. 

  \item \textbf{Ranking Bias.} \textit{The idea that top-ranked results are the most relevant and important will result in attraction of more clicks than others}. This bias affects search engines~\cite{Baeza-Yates:2018:BW:3229066.3209581} and crowdsourcing applications~\cite{lerman2014leveraging}.
 \end{enumerate}
  \item \textbf{Popularity Bias}. \textit{Items that are more popular tend to be exposed more. However, popularity metrics are subject to manipulation---for example, by fake reviews or social bots}  \cite{nematzadeh2017algorithmic}. As an instance, this type of bias can be seen in search engines \cite{nematzadeh2017algorithmic,816269} or recommendation systems where popular objects would be presented more to the public. But this presentation may not be a result of good quality; instead, it may be due to other biased factors.
    \item \textbf{Emergent Bias.} \textit{Emergent bias occurs as a result of use and interaction with real users. This bias arises as a result of change in population, cultural values, or societal knowledge usually some time after the completion of design} \cite{Friedman:1996:BCS:230538.230561}. This type of bias is more likely to be observed in user interfaces, since interfaces tend to reflect the capacities, characteristics, and habits of prospective users by design \cite{Friedman:1996:BCS:230538.230561}. This type of bias can itself be divided into more subtypes, as discussed in detail in \cite{Friedman:1996:BCS:230538.230561}.
    \item \textbf{Evaluation Bias.} \textit{Evaluation bias happens during model evaluation} \cite{suresh2019framework}. This includes the use of inappropriate and disproportionate benchmarks for evaluation of applications such as Adience and IJB-A benchmarks. These benchmarks are used in the  evaluation of facial recognition systems that were biased toward skin color and gender \cite{pmlr-v81-buolamwini18a}, and can serve as examples for this type of bias \cite{suresh2019framework}.
\end{enumerate}
\subsubsection{User to Data}
Many data sources used for training ML models are user-generated. Any inherent biases in users might be reflected in the data they generate. Furthermore, when user behavior is affected/modulated by an algorithm, any biases present in those algorithm might introduce bias in the data generation process. Here we list several important types of such biases.
\begin{enumerate}

     \item  \textbf{Historical Bias.}\textit{ Historical bias is the already existing bias and socio-technical issues in the world and can seep into from the data generation process even given a perfect sampling and feature selection } \cite{suresh2019framework}. An example of  this type of bias can be found in a 2018 image search result where searching for women CEOs ultimately resulted in fewer female CEO images due to the fact that only 5\% of Fortune 500 CEOs were woman---which would cause the search results to be biased towards male CEOs \cite{suresh2019framework}. These search results were of course reflecting the reality, but whether or not the search algorithms should reflect this reality is an issue worth considering.
  
    \item \textbf{Population Bias.} \textit{Population bias arises when statistics, demographics, representatives, and user characteristics are different in the user population 
    of the platform from the original target population}  \cite{olteanu2016social}.
    Population bias creates non-representative data. An example of this type of bias can arise  from different user demographics on different social platforms, such as women being more likely to use Pinterest, Facebook, Instagram, while men being more active in online forums like Reddit or Twitter. More such examples and statistics related to  social media use among young adults according to gender, race, ethnicity, and parental educational background can be found in \cite{10.1111/j.1083-6101.2007.00396.x}.
  
    \item \textbf{Self-Selection Bias.} \textit{Self-selection bias\footnote{\label{fl}https://data36.com/statistical-bias-types-explained/} is a subtype of the  selection or sampling bias in which subjects of the research select themselves.} An example of this type of bias can be observed in an opinion poll to measure enthusiasm for a political candidate, where the most enthusiastic supporters are more likely to complete the poll. 
    
    \item \textbf{Social Bias.} \textit{Social bias happens when others' actions affect our judgment.}  \cite{Baeza-Yates:2018:BW:3229066.3209581}. An example of this type of bias can be a case where we want to rate or review an item with a low score, but when influenced by other high ratings, we change our scoring thinking that perhaps we are being too harsh \cite{wang2014amazon,Baeza-Yates:2018:BW:3229066.3209581}.

    \item \textbf{Behavioral Bias.} \textit{Behavioral bias arises from different user behavior across platforms, contexts, or different datasets} \cite{olteanu2016social}. An example of this type of bias can be observed in \cite{miller2016blissfully}, where authors show how differences in emoji representations among platforms can result in different reactions and behavior from people and sometimes even leading to communication errors.

    \item \textbf{Temporal Bias.} \textit{Temporal bias arises from differences in populations and behaviors over time} \cite{olteanu2016social}. An example can be observed in Twitter where  people talking about a particular topic start using a hashtag at some point to capture attention, then continue the discussion about the event without using the hashtag \cite{tufekci2014big,olteanu2016social}.

    \item \textbf{Content Production Bias.} \textit{Content Production bias arises from structural, lexical, semantic, and syntactic differences in the contents generated by users} \cite{olteanu2016social}. An example of this type of bias can be seen in \cite{2cac1d4b56eb4e43ad55a690a5595a91} where the differences in use of language across different gender and age groups is discussed. The differences in use of language can also be seen across and within countries and populations.
     \end{enumerate}

Existing work tries to categorize these bias definitions into groups, such as definitions falling solely under data or user interaction. However, due to the existence of the feedback loop phenomenon \cite{chouldechova2018frontiers}, 
these definitions are intertwined, and we need a categorization which closely models this situation. This feedback loop is not only existent between the data and the algorithm, but also between the algorithms and user interaction \cite{chaney2018algorithmic}. Inspired by these papers, we modeled categorization of bias definitions, as shown in Figure \ref{fig:cycle}, and grouped these definitions on the arrows of the loop where we thought they were most effective. We emphasize the fact again that these definitions are intertwined, and one should consider how they affect each other in this cycle, and address them accordingly.
\subsection{Data Bias Examples}
There are multiple ways that discriminatory bias can seep into data. 
For instance, using unbalanced data can create biases against underrepresented groups. \cite{zou2018ai} analyzes some examples of the biases that can exist in the data and algorithms and offer some recommendations and suggestions toward mitigating these issues. 
\subsubsection{Examples of Bias in Machine Learning Data}
In \cite{pmlr-v81-buolamwini18a}, the authors show that datasets like IJB-A and Adience are imbalanced and  contain mainly light-skinned subjects---79.6\% in IJB-A and 86.2\% in Adience. This can bias the analysis towards dark-skinned groups who are underrepresented in the data. In another instance, the way we use and analyze our data can create bias when we do not consider different subgroups in the data. In \cite{pmlr-v81-buolamwini18a}, the authors also show that considering only male-female groups is not enough, but there is also a need to use race to further subdivide the gender groups into light-skinned females, light-skinned males, dark-skinned males, and dark-skinned females. It's only in this case that we can clearly observe the bias towards dark-skinned females, as previously dark-skinned males would compromise for dark-skinned females and would hide the underlying bias towards this subgroup. Popular machine-learning datasets that serve as a base for most of the developed algorithms and tools can also be biased---which can be harmful to the downstream applications that are based on these datasets. For instance, ImageNet \cite{russakovsky2015imagenet} and Open Images \cite{krasin2017openimages} are two widely used datasets in machine-learning. In \cite{shankar2017no}, researchers showed that these datasets suffer from representation bias and advocate for the need to incorporate geographic diversity and inclusion while creating such datasets. In addition, authors in~\cite{mehrabi-etal-2021-lawyers} write about the existing representational biases in different knowledge bases that are widely used in Natural Language Processing (NLP) applications for different commonsense reasoning tasks.
\subsubsection{Examples of Data Bias in Medical Applications}
These data biases can be more dangerous in other sensitive applications. For example, in medical domains there are many instances in which the data studied and used are skewed toward certain populations---which can have dangerous consequences for the underrepresented communities.
\cite{doi:10.1056/NEJMsa1507092} showed how exclusion of African-Americans resulted in their misclassification in clinical studies, so they became advocates for sequencing the genomes of diverse populations in the data to prevent harm to  underrepresented populations. Authors in \cite{shawfurther} studied the 23andMe genotype dataset and found that out of 2,399 individuals, who have openly shared their genotypes in public repositories, 2,098 (87\%) are European, while only 58 (2\%) are Asian and 50 (2\%) African. Other such studies were conducted in \cite{10.1093/aje/kwx246} which states that UK Biobank, a large and widely used genetic dataset, may not represent the sampling population. Researchers found evidence of a ``healthy volunteer'' selection bias. \cite{vickers2014enhancing} has other examples of studies on existing biases in the data used in the medical domain. \cite{article} also looks at machine-learning algorithms and data utilized in medical fields, and writes about how artificial intelligence in health care has not impacted all patients equally. 
\subsection{Discrimination}
Similar to bias, discrimination is also a source of unfairness. Discrimination can be considered as a source for unfairness that is due to human prejudice and stereotyping based on the sensitive attributes, which may happen intentionally or unintentionally, while bias can be considered as a source for unfairness that is due to the data collection, sampling, and measurement. Although bias can also be seen as a source of unfairness that is due to human prejudice and stereotyping, in the algorithmic fairness literature it is more intuitive to categorize them as such according to the existing research in these areas. In this survey, we mainly focus on concepts that are relevant to algorithmic fairness issues. \cite{romei2011multidisciplinary,marshall1974economics,willborn1984disparate} contain more broad information on discrimination theory that involve more multidisciplinary concepts from legal theory, economics, and social sciences which can be referenced by the interested readers.
\subsubsection{Explainable Discrimination}
Differences in treatment and outcomes amongst different groups can be justified and explained via some attributes in some cases. In situations where these differences are justified and explained, it is not considered to be illegal discrimination and hence called explainable \cite{Kamiran2013}. For instance, authors in \cite{Kamiran2013} state that in the UCI Adult dataset \cite{Asuncion+Newman:2007}, a widely used dataset in the fairness domain, males on average have a higher annual income than females. However, this is because on average females work fewer hours than males per week. Work hours per week is an attribute that can be used to explain low income which needs to be considered. If we make decisions, without considering working hours, such that males and females end up averaging the same income, we will lead to reverse discrimination since we would cause male employees to get lower salary than females. Therefore, explainable discrimination is acceptable and legal as it can be explained through other attributes like working hours. In \cite{Kamiran2013}, authors present a methodology to quantify the explainable and illegal discrimination in data. They argue that methods that do not take the explainable part of the discrimination into account may result in non-desirable outcomes, so they introduce a \textit{reverse} discrimination which is equally harmful and undesirable. 
They explain how to quantify and measure discrimination in data or a classifier's decisions which directly considers illegal and explainable discrimination.
\subsubsection{Unexplainable Discrimination}
In contrast to explainable discrimination, there is unexplainable discrimination in which the discrimination toward a group is unjustified and therefore considered illegal. Authors in \cite{Kamiran2013} also present local techniques for removing only the illegal or unexplainable discrimination, allowing only for explainable differences in decisions. These are preprocessing techniques that change the training data such that it contains no unexplainable discrimination. We expect classifiers trained on this preprocessed data to not capture illegal or unexplainable discrimination. Unexplainable discrimination consists of  \textit{direct} and \textit{indirect}  discrimination.
\begin{enumerate}
\item{\textbf{Direct Discrimination.}}
Direct discrimination happens when protected attributes of individuals explicitly result in non-favorable outcomes toward them \cite{ijcai2017-549}. Typically, there are some traits identified by law on which it is illegal to discriminate against, and it is usually these traits that are considered to be ``protected'' or ``sensitive'' attributes in computer science literature. A list of some of these protected attributes is provided in Table \ref{electionexample} as specified in the Fair Housing and Equal Credit Opportunity Acts (FHA and ECOA) \cite{chen2019fairness}.

\item{\textbf{Indirect Discrimination.}}
In indirect discrimination, individuals appear to be treated based on seemingly neutral and non-protected attributes; however, protected groups, or individuals still get to be treated unjustly as a result of implicit effects from their protected attributes (e.g., the residential zip code of a person can be used in decision making processes such as loan applications. However, this can still lead to racial discrimination, such as redlining, as despite the fact that zip code appears to be a non-sensitive attribute, it may correlate with race because of the population of residential areas.) \cite{ijcai2017-549,rice1996race}.
\end{enumerate}
\subsubsection{Sources of Discrimination}
\begin{enumerate}
\item{\textbf{Systemic Discrimination.}}
Systemic discrimination refers to policies, customs, or behaviors that are a part of the culture or structure of an organization that may perpetuate discrimination against certain subgroups of the population \cite{unitedequal}. 
\cite{rivera2012hiring} found that employers overwhelmingly preferred competent candidates that were culturally similar to them, and shared similar experiences and hobbies. If the decision-makers happen to belong overwhelmingly to certain subgroups, this may result in discrimination against competent candidates that do not belong to these subgroups. 

\item{\textbf{Statistical Discrimination.}}
Statistical discrimination is a phenomenon where decision-makers use average group statistics to judge an individual belonging to that group. It usually occurs when the decision-makers (e.g., employers, or law enforcement officers) use an individual's obvious, recognizable characteristics as a proxy for either hidden or more-difficult-to-determine characteristics, that may actually be relevant to the outcome \cite{phelps1972statistical}. 
\end{enumerate}

\section{Algorithmic Fairness}
\label{sec:alg-fairness}
Fighting against bias and discrimination has a long history in philosophy and psychology, and recently in machine-learning. However, in order to be able to fight against discrimination and achieve fairness, one should first define fairness. Philosophy and psychology have tried to define the concept of fairness long before computer science. The fact that no universal definition of fairness exists shows the difficulty of solving this problem \cite{Saxena:2019:PF:3306618.3314314}. Different preferences and outlooks in different cultures lend a preference to different ways of looking at fairness, which makes it harder to come up with just a single definition that is acceptable to everyone in a situation. Indeed, even in computer science, where most of the work on proposing new fairness constraints for algorithms has come from the West, and a lot of these papers use the same datasets and problems to show how their constraints perform, there is still no clear agreement on which constraints are the most appropriate for those problems. Broadly, fairness is the absence of any prejudice or favoritism towards an individual or a group based on their intrinsic or acquired traits in the context of decision-making \cite{saxena2019fairness}. Even though fairness is an incredibly desirable quality in society, it can be surprisingly difficult to achieve in practice. With these challenges in mind, many fairness definitions are proposed to address different algorithmic bias and discrimination issues discussed in the previous section.

\subsection{Definitions of Fairness }
In \cite{binns2018fairness}, authors studied fairness definitions in political philosophy and tried to tie them to machine-learning. Authors in \cite{hutchinson201950} studied the 50-year history of fairness definitions in the areas of education and machine-learning. In \cite{verma2018fairness}, authors listed and explained some of the definitions used for fairness in algorithmic classification problems. In \cite{saxena2019fairness}, authors studied the general public's perception of some of these fairness definitions in computer science literature. Here we will reiterate and provide some of the most widely used definitions, along with their explanations inspired from \cite{verma2018fairness}.\\
\\
\textbf{Definition 1.} \textit{(Equalized Odds). The definition of equalized odds, provided by \cite{hardt2016equality}, states that ``A predictor \^{Y} satisfies equalized odds with respect to protected attribute A and outcome Y, if \^{Y} and A are independent conditional on Y. P(\^{Y}=1|A=0,Y =y) = P(\^{Y}=1|A=1,Y =y) , y$\in$\{0,1\}''}. This means that the probability of a person in the positive class being correctly assigned a positive outcome and the probability of a person in a negative class being incorrectly assigned a positive outcome should both be the same for the protected and unprotected group members \cite{verma2018fairness}. In other words, the equalized odds definition states that the protected and unprotected groups should have equal rates for true positives and false positives.
\\
\\
\textbf{Definition 2.} \textit{(Equal Opportunity). ``A binary predictor \^{Y} satisfies equal opportunity with respect to A and Y if P(\^{Y}=1|A=0,Y=1) = P(\^{Y}=1|A=1,Y=1)''}
\cite{hardt2016equality}. This means that the probability of a person in a positive class being assigned to a positive outcome should be equal for both protected and unprotected (female and male) group members \cite{verma2018fairness}. In other words, the equal opportunity definition states that the protected and unprotected groups should have equal true positive rates.
\\
\\
\textbf{Definition 3.} \textit{(Demographic Parity). Also known as statistical parity. ``A predictor \^{Y} satisfies demographic parity if P(\^{Y} |A = 0) = P(\^{Y}|A = 1)''} \cite{NIPS2017_6995,Dwork:2012:FTA:2090236.2090255}. The likelihood of a positive outcome \cite{verma2018fairness} should be the same regardless of whether the person is in the protected (e.g., female) group.
\\
\\
\textbf{Definition 4.} \textit{(Fairness Through Awareness). ``An algorithm is fair if it gives similar predictions to similar individuals''} \cite{NIPS2017_6995,Dwork:2012:FTA:2090236.2090255}. In other words, any two individuals who are similar with respect to a similarity (inverse distance) metric defined for a particular task should receive a similar outcome.\\
\\
\textbf{Definition 5.} \textit{(Fairness Through Unawareness). ``An algorithm is fair as long as any protected attributes A are not explicitly used in the decision-making process''} \cite{NIPS2017_6995,grgic2016case}.\\
\\
\textbf{Definition 6.} \textit{(Treatment Equality).  ``Treatment equality is achieved when the ratio of false negatives and false positives is the same for both protected group categories''} \cite{berk2018fairness}.  \\
\\
\textbf{Definition 7.} \textit{(Test Fairness). ``A score S = S(x) is test fair (well-calibrated) if it reflects the same likelihood of recidivism irrespective of the individual's group membership, R. That is, if for all values of s,
P(Y =1|S=s,R=b)=P(Y =1|S=s,R=w)''} \cite{chouldechova2017fair}. In other words, the test fairness definition states that for any predicted probability score S, people in both protected and unprotected groups must have equal probability of correctly belonging to the positive class \cite{verma2018fairness}.\\
\\
\textbf{Definition 8.} \textit{(Counterfactual Fairness). ``Predictor \^{Y} is counterfactually fair if under any context X =x and A=a,
P($\hat{Y}_{A\xleftarrow{}a }$(U)=y|X =x,A=a)=P($\hat{Y}_{A\xleftarrow{}a'}$(U)=y|X =x,A=a), (for all y and for any value $a'$ attainable by A''} \cite{NIPS2017_6995}. The counterfactual fairness definition is based on the ``intuition that a decision is fair towards an individual if it is the same in both the actual world and a counterfactual world where the individual belonged to a different demographic group.''
\\
\\
\textbf{Definition 9.} \textit{(Fairness in Relational Domains). ``A notion of fairness that is able to capture the relational structure in a domain---not only by taking attributes of individuals into consideration but by taking into account the social, organizational, and other connections between individuals''} \cite{Farnadi:2018:FRD:3278721.3278733}.\\
\\
\textbf{Definition 10.} \textit{(Conditional Statistical Parity). For a set of legitimate factors L, predictor \^{Y} satisfies conditional statistical parity if P(\^{Y} |L=1,A = 0) = P(\^{Y}|L=1,A = 1)} \cite{corbett2017algorithmic}. Conditional statistical parity states that people in both protected and unprotected (female and male) groups should have equal probability of being assigned to a positive outcome given a set of legitimate factors L \cite{verma2018fairness}.
\\
\\
Fairness definitions fall under different types as follows:
\begin{enumerate}
\item{\textbf{Individual Fairness.} Give similar predictions to similar individuals \cite{NIPS2017_6995, Dwork:2012:FTA:2090236.2090255}}.
\item{\textbf{Group Fairness.} Treat different groups equally \cite{NIPS2017_6995,Dwork:2012:FTA:2090236.2090255}}.
\item{\textbf{Subgroup Fairness}. Subgroup fairness intends to obtain the best properties of the group and individual notions of fairness. It is different than these notions but uses them in order to obtain better outcomes. It picks a group fairness constraint like equalizing false positive and asks whether this constraint holds over a large collection of subgroups \cite{kearns2018preventing,kearns2019empirical}}.
\end{enumerate}
\begin{table}[t]
\centering
\begin{tabular}{ |p{5cm}||p{2cm}|p{1.5cm}|p{1.5cm}|p{1.5cm}|}
 \hline
Name & Reference &Group & Subgroup &Individual\\
 \hline
Demographic parity&\cite{NIPS2017_6995}\cite{Dwork:2012:FTA:2090236.2090255}&\checkmark && \\[0.5pt]
 \hline
 Conditional statistical parity&\cite{corbett2017algorithmic}&\checkmark&&\\[0.5pt]
 \hline
 Equalized odds &\cite{hardt2016equality}&\checkmark&&\\[0.5pt]
  \hline
  Equal opportunity&\cite{hardt2016equality}&\checkmark&&\\[0.5pt]
  \hline
  Treatment equality&\cite{berk2018fairness}&\checkmark&&\\[0.5pt]
  \hline
  Test fairness&\cite{chouldechova2017fair}&\checkmark&&\\[0.5pt]
  \hline
  Subgroup fairness & \cite{kearns2018preventing}\cite{kearns2019empirical}&& \checkmark&\\[0.5pt]
  \hline
  Fairness  through  unawareness&\cite{NIPS2017_6995}\cite{grgic2016case}&&&\checkmark\\[0.5pt]
  \hline
  Fairness  through  awareness&\cite{Dwork:2012:FTA:2090236.2090255}&&&\checkmark\\[0.5pt]
   \hline
   Counterfactual fairness&\cite{NIPS2017_6995}&&&\checkmark\\[0.5pt]
   \hline
\end{tabular}
\caption{Categorizing different fairness notions into group, subgroup, and individual types.}
\label{fairtypes}
\end{table}
It is important to note that according to \cite{kleinberg2016inherent}, it is impossible to satisfy some of the fairness constraints at once except in highly constrained special cases. 
In \cite{kleinberg2016inherent}, the authors show the inherent incompatibility of two conditions: calibration and balancing the positive and negative classes. These cannot be satisfied simultaneously with each other unless under certain constraints; therefore, it is important to take the context and application in which fairness definitions need to be used into consideration and use them accordingly \cite{selbst2019fairness}.
Another important aspect to consider is time and temporal analysis of the impacts that these definitions may have on individuals or groups. In \cite{liu2018delayed} authors show that current fairness definitions are not always helpful and do not promote improvement for sensitive groups---and can actually be harmful when analyzed over time in some cases. They also show that measurement errors can also act in favor of these fairness definitions; therefore, they show how temporal modeling and measurement are important in evaluation of fairness criteria and introduce a new range of trade-offs and challenges toward this direction. It is also important to pay attention to the sources of bias and their types when trying to solve fairness-related questions.\\

\section{Methods for Fair Machine Learning}
\label{sec:methods}
There have been numerous attempts to address bias in artificial intelligence in order to achieve fairness; these stem from domains of AI. In this section we will enumerate different domains of AI, and the work that has been produced by each community to combat bias and unfairness in their methods. Table~\ref{domain} provides an overview of the different areas that we focus upon in this survey. 

While this section is largely domain-specific, it can be useful to take a cross-domain view. Generally, methods that target biases in the algorithms fall under three categories: 
\begin{enumerate}
    \item \textbf{Pre-processing.} Pre-processing techniques try to transform the data so that the underlying discrimination is removed \cite{d2017conscientious}. If the algorithm is allowed to modify the training data, then pre-processing can be used \cite{bellamy2018ai}. 
    \item \textbf{In-processing.} In-processing techniques try to modify and change state-of-the-art learning algorithms in order to remove discrimination during the model training process \cite{d2017conscientious}. If it is allowed to change the learning procedure for a machine learning model, then in-processing can be used during the training of a model--- either by incorporating changes into the objective function or imposing a constraint \cite{bellamy2018ai,berk2017convex}.
    \item \textbf{Post-processing.} Post-processing is performed after training by accessing a holdout set which was not involved during the training of the model \cite{d2017conscientious}. If the algorithm can only treat the learned model as a black box without any ability to modify the training data or learning algorithm, then only post-processing can be used in which the labels assigned by the black-box model initially get reassigned based on a function during the post-processing phase \cite{bellamy2018ai,berk2017convex}.
\end{enumerate}
Examples of some existing work and their categorization into these types is shown in Table \ref{prepost}.
These methods are not just limited to general machine learning techniques, but because  of 
AI's  popularity, they have expanded to different domains such as natural language processing and deep learning. From learning fair representations \cite{creager2019flexibly,louizos2016variational,moyer2018invariant} to learning fair word embeddings \cite{zhao2018learning,bolukbasi2016man,gonen2019lipstick}, debiasing methods have been proposed in different AI applications and domains.
Most of these methods try to avoid unethical interference of sensitive or protected attributes into the decision-making process, while others target exclusion bias by trying to include users from sensitive groups. In addition, some works try to satisfy one or more of the fairness notions in their methods, such as disparate learning processes (DLPs) which try to satisfy notions of treatment disparity and impact disparity by allowing the protected attributes  during the training phase but avoiding them during prediction time \cite{lipton2017does}. A list of protected or sensitive attributes is provided in Table \ref{electionexample}. They point out what attributes should not affect the outcome of the decision in housing loan or credit card decision-making \cite{chen2019fairness} according to the law. Some of the existing work tries to treat sensitive attributes as noise to disregard their effect on decision-making, while some causal methods use causal graphs, and disregard some paths in the causal graph that result in sensitive attributes affecting the outcome of the decision. Different bias-mitigating methods and techniques are discussed below for different domains---each targeting a different problem in different areas of machine learning in detail. This can expand the horizon of the reader on where and how bias can affect the system and try to help researchers  carefully look at various new problems concerning  potential places where discrimination and bias can affect the outcome of a system. 
\subsection{Unbiasing Data}
Every dataset is the result of several design decisions made by the data curator. Those decisions have consequences for the fairness of the resulting dataset, which in turn affects the resulting algorithms. In order to mitigate the effects of bias in data, some general methods have been proposed that advocate  having good practices while using data, such as having datasheets that would act like a supporting document for the data reporting the dataset creation method, its characteristics, motivations, and its skews \cite{gebrudatasheets,benjamintowards}. \cite{bender-friedman-2018-data} proposes a similar approach for the NLP applications. A similar suggestion has been proposed for models in \cite{Mitchell:2019:MCM:3287560.3287596}. Authors in \cite{holland2018dataset} also propose having labels, just like nutrition labels on food, in order to better categorize each data for each task. In addition to these general techniques, some work has targeted more specific types of biases. For example, \cite{kievit2013simpson} has proposed methods to test for cases of Simpson's paradox in the data, and \cite{alipourfard2018wsdm,alipourfard2018icwsm} proposed methods to discover Simpson's paradoxes in data automatically. Causal models and graphs were also used in some work to detect direct discrimination in the data along with its prevention technique that modifies the data such that the predictions would be absent from direct discrimination \cite{zhang2017achieving}. \cite{6175897} also worked on  preventing discrimination in data mining, targeting direct, indirect, and simultaneous effects. Other pre-processing approaches, such as messaging \cite{4909197}, preferential sampling \cite{Kamiran2012,kamiran2010classification}, disparate impact removal \cite{10.1145/2783258.2783311}, also aim to remove biases from the data.
\begin{table}[h]
\centering
{\begin{tabular}{ |p{4cm}||p{7cm}|}
 \hline
  \textbf{Area}&\textbf{Reference(s)}\\
 \hline
 Classification&\cite{kamishima2012fairness} \cite{pmlr-v81-menon18a} \cite{goel2018non} \cite{Krasanakis:2018:ASR:3178876.3186133} \cite{pmlr-v97-ustun19a} \cite{hardt2016equality} \cite{zafar2015fairness} \cite{woodworth2017learning} \cite{huang2019stable} \cite{calders2010three} \cite{wu2018fairnessaware} \cite{oneto2019taking} \cite{pmlr-v81-dwork18a} \cite{jiangwasserstein} \cite{kamiran2010classification}\\
 \hline
 Regression&\cite{berk2017convex} \cite{agarwal2019fair}\\
 \hline
 PCA&\cite{Samadi:2018:PFP:3327546.3327755}\\
 \hline
  Community detection&\cite{mehrabi2019debiasing}\\
 \hline
 Clustering&\cite{chen2019proportionally} \cite{pmlr-v97-backurs19a}\\
 \hline
  Graph embedding&\cite{bose2019compositional}\\
 \hline
 Causal inference&\cite{loftus2018causal} \cite{ijcai2017-549} \cite{8477109} \cite{Zhang2017} \cite{nabi2018fair} \cite{nabi2018learning} \cite{Zhang:2016:STD:3060832.3061001} \cite{kilbertus2017avoiding} \cite{qureshi2016causal} \cite{10.1007/978-3-319-39931-7_9} \\
 \hline
  Variational auto encoders&\cite{louizos2016variational} \cite{amini2019uncovering} \cite{moyer2018invariant} \cite{creager2019flexibly}\\
  \hline
   Adversarial learning&\cite{lemoine2018mitigating} \cite{xu2018fairgan}\\
  \hline
 Word embedding&\cite{bolukbasi2016man} \cite{zhao2018learning} \cite{gonen2019lipstick}  \cite{pmlr-v97-brunet19a} \cite{zhao2019gender} \\
 \hline
 Coreference resolution&\cite{zhao2018gender} \cite{rudinger-etal-2018-gender}\\
 \hline
Language model&\cite{bordia2019identifying}\\
 \hline
  Sentence embedding&\cite{may2019measuring}\\
 \hline
 Machine translation&\cite{font2019equalizing}\\
 \hline
  Semantic role labeling&\cite{zhao2017men}\\
 \hline
 Named Entity Recognition&\cite{mehrabi2019man} \\
 \hline
\end{tabular}}
\caption{List of papers targeting and talking about bias and fairness in different areas.}
\label{domain}
\end{table}
\begin{table}[h]
\centering
\begin{tabular}{ |p{6cm}||p{2cm}|p{2cm}|}
 \hline
 Attribute& FHA & ECOA\\
 \hline
Race&\checkmark&\checkmark\\[0.5pt]
 \hline
 Color&\checkmark&\checkmark\\[0.5pt]
 \hline
 National origin&\checkmark&\checkmark\\[0.5pt]
  \hline
  Religion&\checkmark&\checkmark\\[0.5pt]
  \hline
   Sex&\checkmark&\checkmark\\[0.5pt]
  \hline
   Familial status&\checkmark&\\[0.5pt]
  \hline
   Disability&\checkmark&\\[0.5pt]
  \hline
   Exercised rights under CCPA&&\checkmark\\[0.5pt]
  \hline
   Marital status&&\checkmark\\[0.5pt]
  \hline
   Recipient of public assistance&&\checkmark\\[0.5pt]
  \hline
  Age&&\checkmark\\[0.5pt]
  \hline
\end{tabular}
\caption{A list of the protected attributes as specified in the Fair Housing and Equal Credit Opportunity Acts (FHA and ECOA), from \cite{chen2019fairness}.}
\label{electionexample}
\end{table}
\subsection{Fair Machine Learning}
To address this issue, a variety of methods have been proposed that satisfy some of the fairness definitions or other new definitions depending on the application.
\subsubsection{Fair Classification}
Since classification is a canonical task in machine learning and is widely used in different areas that can be in direct contact with humans, it is important that these types of methods be fair and be absent from biases that can harm some populations. Therefore, certain methods have been proposed \cite{kamishima2012fairness,pmlr-v81-menon18a,goel2018non,Krasanakis:2018:ASR:3178876.3186133} that satisfy certain definitions of fairness in classification. For instance, in \cite{pmlr-v97-ustun19a} authors try to satisfy subgroup fairness in classification, equality of opportunity and equalized odds in \cite{hardt2016equality}, both disparate treatment and disparate impact in \cite{zafar2015fairness,aghaei2019learning}, and equalized odds in \cite{woodworth2017learning}. Other methods try to not only satisfy some fairness constraints but to also be stable toward change in the test set \cite{huang2019stable}. The authors in \cite{wu2018fairnessaware}, propose a general framework for learning fair classifiers. This framework can be used for formulating fairness-aware classification with fairness guarantees. In another work \cite{calders2010three}, authors propose three different modifications to the existing Naive Bayes classifier for discrimination-free classification. \cite{oneto2019taking} takes a new approach into fair classification by imposing fairness constraints into a Multitask learning (MTL) framework. In addition to imposing fairness during training, this approach can benefit the minority groups by focusing on maximizing the average accuracy of each group as opposed to maximizing the accuracy as a whole without attention to accuracy across different groups. In a similar work \cite{pmlr-v81-dwork18a}, authors propose a decoupled classification system where a separate classifier is learned for each group. They use transfer learning to reduce the issue of having less data for minority groups. In \cite{jiangwasserstein} authors propose to achieve fair classification by mitigating the dependence of the classification outcome on the sensitive attributes by utilizing the Wasserstein distance measure. In \cite{kamiran2010classification} authors propose the Preferential Sampling (PS) method to create a discrimination free train data set. They then learn a classifier on this discrimination free dataset to have a classifier with no discrimination. In~\cite{mehrabi2021attributing}, authors propose a post-processing bias mitigation strategy that utilizes attention mechanism for classification and that can provide interpretability.
\begin{table}[h]
\centering
\begin{tabular}{ |p{3.6cm}||p{1.5cm}|p{2.5cm}|p{2.5cm}|p{2.5cm}|}
 \hline
 Algorithm&Reference&Pre-Processing&In-Processing& Post-Processing\\
 \hline
  Community detection&\cite{mehrabi2019debiasing}&\checkmark&&\\[0.5pt]
  \hline
   Word embedding&\cite{pmlr-v97-brunet19a}&\checkmark&&\\[0.5pt]
  \hline
   Optimized pre-processing&\cite{NIPS2017_6988}&\checkmark&&\\[0.5pt]
  \hline
   Data pre-processing&\cite{Kamiran2012}&\checkmark&&\\[0.5pt]
  \hline
 Classification&\cite{zafar2015fairness}& &\checkmark&\\[0.5pt]
 \hline
 Regression&\cite{berk2017convex}&&\checkmark&\\[0.5pt]
 \hline
 Classification&\cite{kamishima2012fairness}&&\checkmark&\\[0.5pt]
  \hline
  Classification&\cite{wu2018fairnessaware}&&\checkmark&\\[0.5pt]
  \hline
  Adversarial learning&\cite{lemoine2018mitigating}&&\checkmark&\\[0.5pt]
  \hline
  Classification&\cite{hardt2016equality}&&&\checkmark\\[0.5pt]
  \hline
  Word embedding&\cite{bolukbasi2016man}&&&\checkmark\\[0.5pt]
  \hline
  Classification&\cite{NIPS2017_7151}&&&\checkmark\\[0.5pt]
  \hline
   Classification&\cite{mehrabi2021attributing}&&&\checkmark\\[0.5pt]
  \hline
\end{tabular}
\caption{Algorithms categorized into their appropriate groups based on being pre-processing, in-processing, or post-processing.}
\label{prepost}
\end{table}
\subsubsection{Fair Regression}
\cite{berk2017convex} proposes a fair regression method along with evaluating it with a measure introduced as the ``price of fairness'' (POF) to measure accuracy-fairness trade-offs. They introduce three fairness penalties as follows:\\

Individual Fairness: The definition for individual fairness as stated in~\cite{berk2017convex}, ``for every cross pair $(x, y)\in S_{1}$, $(x', y')\in S_{2}$, a model $w$ is penalized for how differently it treats $x$ and $x'$ (weighted by a function of $|y - y'|$) where $S_{1}$ and $S_{2}$ are different groups from the sampled population.'' Formally, this is operationalized as
\[f_{1}(w,S)= \frac{1}{n_{1}n_{2}} \sum_{\substack{(x_{i},y_{i})\in S_{1}\\(x_{j},y_{j})\in S_{2}}}d(y_{i},y_{j})(w.x_{i}-w.x_{j})^{2} \]
Group Fairness: "On average, the two groups' instances should have similar labels (weighted by the nearness of the labels of the instances)" \cite{berk2017convex}.
\[f_{2}(w,S)= \Bigg(\frac{1}{n_{1}n_{2}} \sum_{\substack{(x_{i},y_{i})\in S_{1}\\(x_{j},y_{j})\in S_{2}}}d(y_{i},y_{j})(w.x_{i}-w.x_{j})\Bigg)^{2} \]
Hybrid Fairness: "Hybrid fairness requires both positive and both negatively labeled cross pairs to be treated similarly in an average over the two groups" \cite{berk2017convex}.
\[f_{3}(w,S)= \Bigg ( \sum_{\substack{(x_{i},y_{i})\in S_{1}\\(x_{j},y_{j})\in S_{2}\\y_{i}=y_{j}=1}} \frac{d(y_{i},y_{j})(w.x_{i}-w.x_{j})}{n_{1,1}n_{2,1}} \Bigg)^{2} + \Bigg ( \sum_{\substack{(x_{i},y_{i})\in S_{1}\\(x_{j},y_{j})\in S_{2}\\y_{i}=y_{j}=-1}} \frac{d(y_{i},y_{j})(w.x_{i}-w.x_{j})}{n_{1,-1}n_{2,-1}} \Bigg)^{2} \]
In addition to the previous work, \cite{agarwal2019fair} considers the fair regression problem formulation with regards to two notions of fairness statistical (demographic) parity and bounded group loss. \cite{aghaei2019learning} uses decision trees to satisfy disparate impact and treatment in regression tasks in addition to classification.
\subsubsection{Structured Prediction}
In \cite{zhao2017men}, authors studied the semantic role-labeling models and a famous dataset, imSitu, and realized that only 33\% of agent roles in cooking images are man, and the rest of 67\% cooking images have woman as agents in the imSitu training set. They also noticed that in addition to the existing bias in the dataset, the model would amplify the bias such that after training a model\footnote{Specifically, a Conditional Random Field (CRF)} on the dataset, bias is magnified for ``man'', filling only 16\% of cooking images. Under these observations, the authors of the paper \cite{zhao2017men} show that structured prediction models have the risk of leveraging social bias. Therefore, they propose a calibration algorithm called RBA (reducing bias amplification); RBA is a technique for debiasing models by calibrating prediction in structured prediction. The idea behind RBA is to ensure that the model predictions follow the same distribution in the training data. They study two cases: multi-label object and visual semantic role labeling classification. They show how these methods amplify the existing bias in data. 
\subsubsection{Fair PCA}
In \cite{Samadi:2018:PFP:3327546.3327755} authors show that vanilla PCA can exaggerate the error in reconstruction in one group of people over a different group of equal size, so they propose a fair method to create representations with similar richness for different populations---not to make them indistinguishable, or to hide dependence on a sensitive or protected attribute. They show that vanilla PCA on the labeled faces in the wild (LFW) dataset~\cite{huang2008labeled} has a lower reconstruction error rate for men than for women faces, even if the sampling is done with an equal weight for both genders. They intend to introduce a dimensionality reduction technique which maintains similar fidelity for different groups and populations in the dataset. Therefore, they introduce Fair PCA and define a fair dimensionality reduction algorithm. Their definition of Fair PCA (as an optimization function) is as follows, in which $A$ and $B$ denote two subgroups, $U_A$ and $U_B$ denote matrices whose rows correspond to rows of $U$ that contain members of subgroups $A$ and $B$ given $m$ data points in $R^n$:\\
\[ min_{U \in R^{m \times n}, rank(U) \leq d} \; max \Bigg \{ \frac{1}{|A|} loss(A , U_{A}) ,  \frac{1}{|B|} loss(B , U_{B})   \Bigg \} \]
And their proposed algorithm is a two-step process listed below:
\begin{enumerate}
\item{Relax the Fair PCA objective to a semidefinite program (SDP) and solve it.}
\item{Solve a linear program that would reduce the rank of the solution.}
\end{enumerate}
\subsubsection{Community Detection/Graph Embedding/Clustering}
Inequalities in online communities and social networks can also potentially be another place where bias and discrimination can affect the populations. For example, in online communities users with a fewer number of friends or followers face a disadvantage of being heard in online social media \cite{mehrabi2019debiasing}. In addition, existing methods, such as community detection methods, can amplify this bias by ignoring these low-connected users in the network or by wrongfully  assigning them to the irrelevant and small communities. In \cite{mehrabi2019debiasing} authors show how this type of bias exists and is  perpetuated by the existing community detection methods. They propose a new attributed community detection method, called CLAN, to mitigate the harm toward disadvantaged groups in online social communities. CLAN is a two-step process that considers the network structure alongside node attributes to address exclusion bias, as indicated below:
\begin{enumerate}
\item{Detect communities using modularity values (Step 1-unsupervised using only network structure).}
\item{Train a classifier to classify users in the minor groups, putting them into one of the major groups using held-out node attributes (Step 2-supervised using other node attributes).}
\end{enumerate}
Fair methods in domains similar to community detection are also proposed, such as graph embedding \cite{bose2019compositional} and clustering \cite{chen2019proportionally,pmlr-v97-backurs19a}. 
\subsubsection{Causal Approach to Fairness}
Causal models can ascertain causal relationships between variables. Using causal graphs one can represent these causal relationships between variables (nodes of the graph) through the edges of the graph. These models can be used to remove unwanted causal dependence of outcomes on sensitive attributes such as gender or race in designing systems or policies \cite{loftus2018causal}. Many researchers have used causal models and graphs to solve  fairness-related concerns in machine learning. In \cite{loftus2018causal,chiappa19causal}, authors discuss in detail the subject of  causality and its importance while designing fair algorithms. There has been much research on discrimination discovery and removal that uses causal models and graphs in order to make decisions that are irrespective of sensitive attributes of groups or individuals. For instance, in \cite{ijcai2017-549} authors propose a causal-based framework that detects direct and indirect discrimination in the data along with their removal techniques. \cite{8477109} is an extension to the previous work. \cite{Zhang2017} gives a nice overview of most of the previous work done in this area by the authors, along with discussing  system-, group-, and individual-level discrimination and solving each using their previous methods,  in addition to targeting direct and indirect discrimination. By expanding on the previous work and generalizing it, authors in \cite{nabi2018fair} propose a similar pathway approach for fair inference using causal graphs; this would restrict certain problematic and discriminative pathways in the causal graph flexibly given any set of constraints. This holds when the path-specific effects can be identified from the observed distribution. In \cite{chiappa19path} authors introduce the path-specific counterfactual fairness definition which is an extension to counterfactual fairness definition \cite{NIPS2017_6995} and propose a method to achieve it further extending the work in \cite{nabi2018fair}. In \cite{nabi2018learning} authors extended a formalization of algorithmic fairness from their previous work to the setting of learning optimal policies that are subject to constraints based on definitions of fairness. They describe several strategies for learning optimal policies by modifying some of the existing strategies, such as Q-learning, value search, and G-estimation, based on some fairness considerations. In \cite{Zhang:2016:STD:3060832.3061001} authors only target discrimination discovery and no removal by finding instances similar to another instance and observing if a  change in the protected attribute will change the outcome of the decision. If so, they declare the existence of discrimination. In \cite{kilbertus2017avoiding}, authors define the following two notions of discrimination---unresolved discrimination and proxy discrimination---as follows: \\
\textbf{Unresolved Discrimination:} "A variable V in a causal graph exhibits unresolved discrimination if there exists a directed path from A to V that is not blocked by a resolving variable, and V itself is non-resolving" \cite{kilbertus2017avoiding}. \\
\textbf{Proxy Discrimination:} "A variable V in a causal graph exhibits potential proxy discrimination, if there exists a directed path from A to V that is blocked by a proxy variable and V itself is not a proxy" \cite{kilbertus2017avoiding}.
They proposed methods to prevent and avoid them. They also show that no observational criterion can determine whether a predictor exhibits unresolved discrimination; therefore, a causal reasoning framework needs to be incorporated. \\
In \cite{qureshi2016causal}, Instead of using the usual risk difference $RD=p_{1}-p_{2}$, authors propose a causal risk difference $RD^{c}=p_{1}-p_{2}^{c}$ for causal discrimination discovery.
They define $p_{2}^{c}$ to be:\\
\[ p_{2}^{c} = \frac{\sum_{\textbf{s} \in S, dec(\textbf{s})=\ominus}w(\textbf{s})}{\sum_{\textbf{s} \in S}w(\textbf{s})} \]
$RD^{c}$ not close to zero means that there is a bias in decision value due to group 
membership (causal discrimination) or to covariates that have not been accounted 
for in the analysis (omitted variable bias).
This $RD^{c}$ then becomes their causal discrimination measure for discrimination discovery. \cite{10.1007/978-3-319-39931-7_9} is another work of this type that uses causal networks for discrimination discovery.

\subsection{Fair Representation Learning} 
\subsubsection{Variational Auto Encoders}
Learning fair representations and avoiding the unfair interference of sensitive attributes has been introduced in many different research papers. A well-known example is the Variational Fair Autoencoder introduced in \cite{louizos2016variational}. Here,they treat the sensitive variable as the nuisance variable, so that by removing the information about this variable they will get a fair representation. They use a maximum mean discrepancy regularizer to obtain invariance in the posterior distribution over latent variables. Adding this maximum mean discrepancy (MMD) penalty into the lower bound of their VAE architecture satisfies their proposed model for having the Variational Fair Autoencoder.
Similar work, but  not targeting fairness specifically, has been introduced in \cite{jaiswal2018unsupervised}. In \cite{amini2019uncovering} authors also propose a debiased VAE architecture called DB-VAE which learns sensitive latent variables that can bias the model (e.g., skin tone, gender, etc.) and propose an algorithm on top of this DB-VAE using these latent variables to debias systems like facial detection systems. In \cite{moyer2018invariant} authors model their representation-learning task as an optimization objective that would minimize the loss of the mutual information between the encoding and the sensitive variable. The relaxed version of this assumption is shown in Equation \ref{eq1}. They use this in order to learn fair representation and show that adversarial training is unnecessary and in some cases even counter-productive. In Equation \ref{eq1}, c is the sensitive variable and z the encoding of x.
\begin{equation}
   \mathop{min}_{q} \mathcal{L}(q,x)+\lambda I(z,c)
    \label{eq1}
\end{equation}
In \cite{creager2019flexibly}, authors introduce flexibly fair representation learning by disentanglement that disentangles information from multiple sensitive attributes. Their flexible and fair variational autoencoder is not only flexible with respect to downstream task labels but also flexible with respect to sensitive attributes. They address the demographic parity notion of fairness, which can target multiple sensitive attributes or any subset combination of them. 
\subsubsection{Adversarial Learning}
\begin{figure}[H]
\includegraphics[width=.6\textwidth,trim=10cm 0cm 5cm 1cm,clip=true]{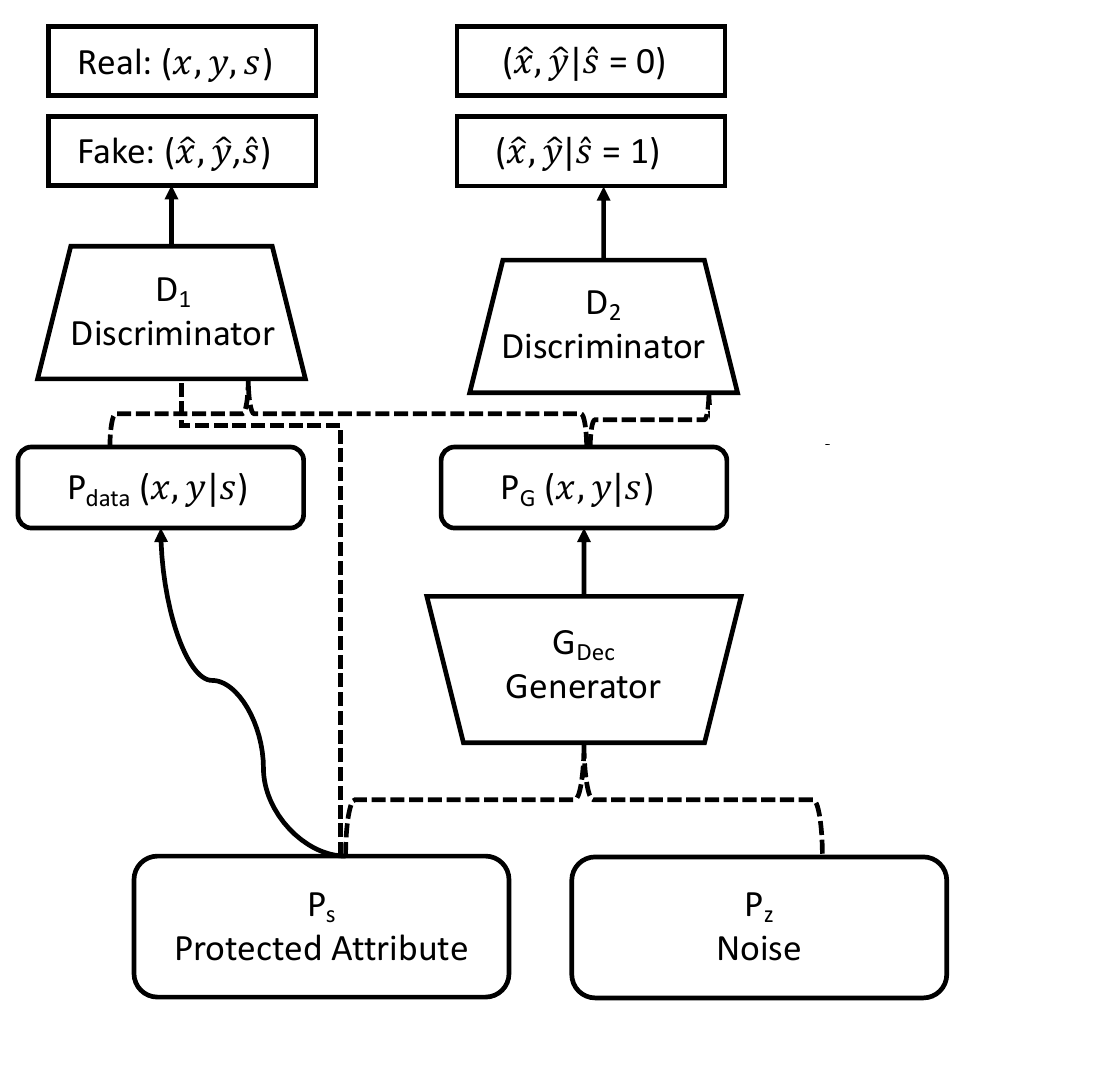}
\caption{Structure of FairGAN as proposed in \cite{xu2018fairgan}.}
\label{adversary}
\end{figure}

\begin{figure}[H]
\includegraphics[width=0.6\textwidth,trim=0cm 12cm 0cm 12cm,clip=true]{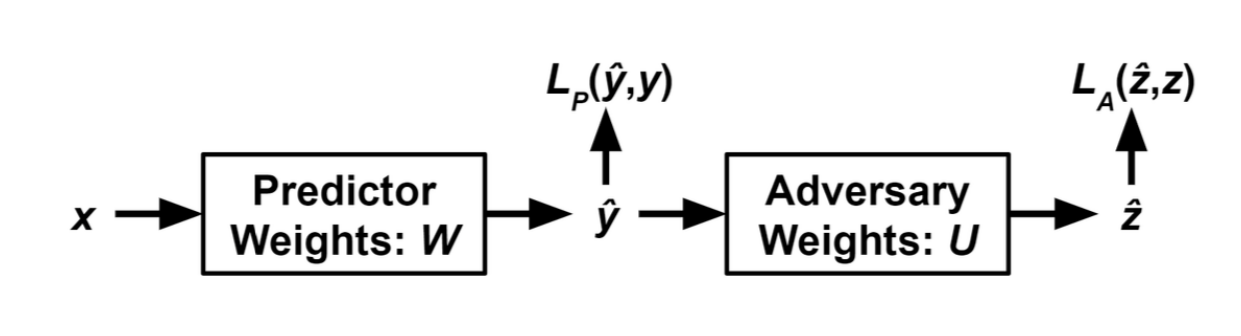}
\caption{The architecture of adversarial network proposed in \cite{lemoine2018mitigating} \textsuperscript{\textcopyright} Brian Hu Zhang.}
\label{adversary1}
\end{figure}

In \cite{lemoine2018mitigating} authors present a framework to mitigate bias in models learned from data with stereotypical associations. They propose a model in which they are trying to maximize accuracy of the predictor on y, and at the same time minimize the ability of the adversary to predict the protected or sensitive variable (stereotyping variable z). The model consists of two parts---the predictor and the adversary---as shown in Figure \ref{adversary1}. In their model, the predictor is trained to predict Y given X. With the help of a gradient-based approach like stochastic gradient descent, the model tries to learn the weights W by minimizing some loss function LP($\hat{y}$, y). The output layer is passed to an adversary, which is another network. This network tries to predict Z. 
The adversary may have different inputs depending on the fairness definition needing to be achieved. For instance, in order to satisfy \textbf{Demographic Parity}, the adversary would try to predict the protected variable Z using only the predicted label $\hat{Y}$ passed as an input to it, while preventing the adversary from learning this is the goal of the predictor. Similarly, to achieve \textbf{Equality of Odds}, the adversary would get the true label Y in addition to the predicted label $\hat{Y}$. To satisfy \textbf{Equality of Opportunity} for a given class y, they would only select instances for the adversary where Y=y. \cite{xu2018fairgan} takes an interesting and different direction toward solving fairness issues using adversarial networks by introducing FairGAN which  generates synthetic data that is free from discrimination and is similar to the real data. They use their newly generated synthetic data from FairGAN, which is now debiased, instead of the real data for training and testing. They do not try to remove discrimination from the dataset, unlike many of the existing approaches, but instead generate new datasets similar to the real one which is debiased and preserves good data utility. The architecture of their FairGAN model is shown in Figure \ref{adversary}. FairGAN consists of two components: a generator $G_{Dec}$ which generates the fake data conditioned on the protected attribute $P_{G}(x,y,s)=P_{G}(x,y|s)P_{G}(s)$ where $P_{G}(s)=P_{data}(s)$, and two discriminators $D_{1}$ and $D_{2}$. $D_{1}$ is trained to differentiate the real data denoted by $P_{data}(x,y,s)$ from the generated fake data denoted by $P_{G}(x,y,s)$. 
In addition to that, for achieving fairness constraints, such as statistical parity, $P_{G}(x,y|s=1)=P_{G}(x,y|s=0)$, the training of $D_{2}$ is such that it emphasizes differentiation of the two types of synthetic (generated by the model) samples $P_{G}(x,y|s=1)$ and $P_{G}(x,y|s=0)$ indicating if the synthetic samples are from the unprotected or protected groups. Here s denotes the protected or the sensitive variable, and we adapted the same notation as in \cite{xu2018fairgan}.
\subsection{Fair NLP} 
\subsubsection{Word Embedding}
In \cite{bolukbasi2016man} authors noticed that while using state-of-the-art word embeddings in word analogy tests, ``man'' would be mapped to ``computer programmer'' and ``woman'' would be mapped to ``homemaker.'' This bias toward woman triggered the authors to propose a method to debias word embeddings by proposing a method that respects the embeddings for gender-specific words but debiases embeddings for gender-neutral words by following these steps: \textit{(Notice that Step 2 has two different options. Depending on whether you target hard debiasing or soft debiasing, you would use either step 2a or 2b)}
\begin{enumerate}
    \item \textbf{Identify gender subspace.} Identifying a direction of the embedding that captures the bias \cite{bolukbasi2016man}.
    \item \textbf{Hard debiasing or soft debiasing:}
    \begin{enumerate}
         \item \textbf{Hard debiasing (neutralize and equalize).} Neutralize puts away the gender subspace from gender-neutral words and makes sure that all the gender-neutral words are removed and zeroed out in the gender subspace \cite{bolukbasi2016man}.
    Equalize makes gender-neutral words to be equidistant from the equality set of gendered words \cite{bolukbasi2016man}.
    \item \textbf{Soft bias correction.} Tries to move as little as possible to retain its similarity to the original embedding as much as possible, while reducing the gender bias. This trade-off is controlled by a parameter \cite{bolukbasi2016man}. 
    \end{enumerate}
\end{enumerate}
Following on the footsteps of these authors, other future work  attempted to tackle this problem \cite{zhao2018learning} by generating a gender-neutral version of (Glove called GN-Glove) that tries to retain gender information in some of the word embedding's learned dimensions, while ensuring that other dimensions are free from this gender effect. This approach primarily relies on Glove as its base model with gender as the protected attribute. However, a recent paper \cite{gonen2019lipstick} argues against these debiasing techniques and states that many recent works on debiasing word embeddings have been superficial, that those techniques just hide the bias and don't actually remove it. A recent work \cite{pmlr-v97-brunet19a} took a new direction and proposed a preprocessing method for the discovery of the problematic documents in the training corpus that have biases in them, and tried to debias the system by perturbing or removing these documents efficiently from the training corpus. In a very  recent work \cite{zhao2019gender}, authors target bias in ELMo's contextualized word vectors and attempt to analyze and mitigate the observed bias in the embeddings. They show that the corpus used for training of ELMo has a significant gender skew, with male entities being nearly three times more common than female entities. This automatically leads to gender bias in these pretrained contextualized embeddings. They propose the following two methods for mitigating the existing bias while using the pretrained embeddings in a downstream task, coreference resolution: 
(1) train-time data augmentation approach, and 
(2) test-time neutralization approach.
\subsubsection{Coreference Resolution}
The \cite{zhao2018gender} paper shows that coreference systems have a gender bias. They introduce a benchmark, called WinoBias, focusing on gender bias in coreference resolution. In addition to that, they introduce a data-augmentation technique that removes bias in the existing state-of-the-art coreferencing methods, in combination with using word2vec debiasing techniques. Their general approach is as follows: They first generate auxiliary datasets using a rule-based approach in which they replace all the male entities with female entities and the other way around. Then they train models with a combination of the original and the auxiliary datasets. They use the above solution in combination with word2vec debiasing techniques to generate word embeddings. They also point out sources of gender bias in coreference systems and propose solutions 
to them. They show that the first source of bias comes from the training data and propose a solution that generates an auxiliary data set by swapping male and female entities. Another case arises from the resource bias (word embeddings are bias), so the proposed solution is to replace Glove with a debiased embedding method. Last,  another source of bias can come from unbalanced gender lists, and balancing the counts in the lists is a solution they proposed. In another work \cite{rudinger-etal-2018-gender}, authors also show the existence of gender bias in three state-of-the-art coreference resolution systems by observing that for many occupations, these systems resolve pronouns in a biased fashion by preferring one gender over the other.
\subsubsection{Language Model}
In \cite{bordia2019identifying} authors introduce a metric for measuring gender bias in a generated text from a language model based on recurrent neural networks that is trained on a text corpus along with measuring the bias in the training text itself. They use Equation \ref{language_model_bias}, where $w$ is any word in the corpus, $f$  is a set of gendered words that belong to the female category, such as she, her, woman, etc., and $m$ to the male category, and measure the bias using the mean absolute and standard deviation of the proposed metric along with fitting a univariate linear regression model over it and then analyzing the effectiveness of each of those metrics while measuring the bias. 
\begin{equation}
    bias(w) = log (\frac{P(w|f)}{P(w|m)})
    \label{language_model_bias}
\end{equation}
In their language model, they also introduce a regularization loss term that would minimize the projection of embeddings trained by the encoder onto the embedding of the gender subspace following the soft debiasing technique introduced in \cite{bolukbasi2016man}. Finally, they evaluate the effectiveness of their method on reducing gender bias and conclude by stating that in order to reduce bias, there is a compromise on perplexity. They also point out the effectiveness of word-level bias metrics over the corpus-level metrics.
\subsubsection{Sentence Encoder}
In \cite{may2019measuring} authors extend the research in detecting bias in word embedding techniques to that of sentence embedding. They try to generalize bias-measuring techniques, such as using the Word Embedding Association Test (WEAT \cite{caliskan2017semantics}) in the context of sentence encoders by introducing their new sentence encoding bias-measuring techniques, the  Sentence Encoder Association Test (SEAT). They used state-of-the-art sentence encoding techniques, such as CBoW, GPT, ELMo, and BERT, and find that although there was  varying evidence of human-like bias in sentence encoders using SEAT, more recent methods like BERT are more immune to biases. That being said, they are not claiming that these models are bias-free, but state that more sophisticated bias discovery techniques may be used in these cases, thereby  encouraging more future work in this area. 
\subsubsection{Machine Translation}
In \cite{font2019equalizing} authors  noticed that when translating the word "friend" in the following two sentences from English to Spanish, they achieved different results---although in both cases this word should be translated the same way.\\
"She works in a hospital, my friend is a nurse."\\
"She works in a hospital, my friend is a doctor."\\
In both of these sentences, "friend" should be translated to the female version of Spanish friend "amiga," but the results were not reflecting this expectation. For the second sentence, friend was translated to "amigo,"---the male version of friend in Spanish. This is because doctor is more stereotypical to males and nurse to females, and the model picks this bias or stereotype and reflects it in its performance. To solve this, authors in \cite{font2019equalizing} build an approach that leverages the fact that machine translation uses word embeddings. They use the existing debiasing methods in word embedding and apply them in the machine translation pipeline. This not only helped them to mitigate the existing bias in their system, but also boosted the performance of their system by one BLUE score. In \cite{prates2018assessing} authors show that Google's translate system can suffer from gender bias by making sentences taken from the U.S. Bureau of Labor Statistics into a dozen languages that are gender neutral, including Yoruba, Hungarian, and Chinese, translating them into English, and showing that Google Translate shows favoritism toward males for stereotypical fields such as STEM jobs. In \cite{vanmassenhove2018getting} authors annotated and analyzed the Europarl dataset \cite{koehn2005europarl}, a large political, multilingual dataset used in machine translation, and discovered that with the exception of the youngest age group (20-30), which represents only a very small percentage of the total amount of sentences (0.71\%), more male data is available in all age groups. They also looked at the entire dataset and showed that 67.39\% of the sentences are produced by male speakers. Furthermore, to mitigate the gender-related issues and to improve morphological agreement in machine translation, they augmented every sentence with a tag on the English source side, identifying the gender of the speaker. This helped the system in most of the cases, but not always, so further work has been suggested for integrating speaker information in other ways.

\subsubsection{Named Entity Recognition} In \cite{mehrabi2019man}, authors investigate a type of existing bias in various named entity recognition (NER) systems. In particular, they observed that in a context where an entity should be tagged as a person entity, such as "John is a person" or "John is going to school", more female names as opposed to male names are being tagged as non-person entities or not being tagged at all. To further formalize their observations, authors propose six different evaluation metrics that would measure amount of bias among different genders in NER systems. They curated templated sentences pertaining to human actions and applied these metrics on names from U.S census data incorporated into the templates. The six introduced measures each aim to demonstrate a certain type of bias and serve a specific purpose in showing various results as follows:
\begin{itemize}
  \item Error Type-1 Unweighted: Through this type of error, authors wanted to recognize the proportion of entities that are tagged as anything other than the person entity in each of the male vs female demographic groups. This could be the entity not being tagged or tagged as other entities, such as location.
  \[ \frac{\sum_{n \in N_f}I(n_{type} \neq PERSON)}{|N_f|}\]
  \item Error Type-1 Weighted: This type of error is similar to its unweighted case except authors considered the frequency or popularity of names so that they could penalize if a more popular name is being tagged wrongfully.
  \[ \frac{\sum_{n \in N_f}freq_f(n_{type} \neq PERSON)}{\sum_{n \in N_f}freq_f(n)},\]
where $freq_f(\cdot)$ indicates the frequency of a name for a particular year in the female census data. Likewise, $freq_m(\cdot)$ indicates the frequency of a name for a particular year in the male census data.
  \item Error Type-2 Unweighted: This is a type of error in which the entity is tagged as other entities, such as location or city. Notice that this error does not count if the entity is not tagged.
  \[ \frac{\sum_{n \in N_f}I(n_{type} \notin \{\emptyset,PERSON\})}{|N_f|},\]
where $\emptyset$ indicates that the name is not tagged.
  \item Error Type-2 Weighted: This error is again similar to its unweighted case except the frequency is taken into consideration.
  \[ \frac{\sum_{n \in N_f}freq_f(n_{type} \notin \{\emptyset,PERSON\})}{\sum_{n \in N_f}freq_f(n)}\]
  \item Error Type-3 Unweighted: This is a type of error in which it reports if the entity is not tagged at all. Notice that even if the entity is tagged as a non-person entity this error type would not consider it.
  \[ \frac{\sum_{n \in N_f}I(n_{type} = \emptyset)}{|N_f|}\]
  \item Error Type-3 Weighted: Again, this error is similar to its unweighted case with frequency taken into consideration.
  \[ \frac{\sum_{n \in N_f}freq_f(n_{type} = \emptyset)}{\sum_{n \in N_f}freq_f(n)}\]
\end{itemize}
Authors also investigate the data that these NER systems are trained on and find that the data is also biased toward female gender by not including as versatile names as there should be to represent female names.

\subsection{Comparison of Different Mitigation Algorithms}
The field of algorithmic fairness is a relatively new area of research and work still needs to be done for its improvement. With that being said, there are already papers that propose fair AI algorithms and bias mitigation techniques and compare different mitigation algorithms using different benchmark datasets in the fairness domain. For instance, authors in \cite{he2020geometric} propose a geometric solution to learn fair representations that removes correlation between protected and unprotected features. The proposed approach can control the trade-off between fairness and accuracy via an adjustable parameter. In this work, authors evaluate the performance of their approach on different benchmark datasets, such as COMPAS, Adult and German, and compare them against various different approaches for fair learning algorithms considering fairness and accuracy measures \cite{jaiswal2018unsupervised,zafar2017fairness,zafar2015fairness,he2020geometric}. In addition, IBM's AI Fairness 360 (AIF360) toolkit \cite{bellamy2018ai} has implemented many of the current fair learning algorithms and has demonstrated some of the results as demos which can be utilized by interested users to compare different methods with regards to different fairness measures.

\section{Challenges and Opportunities for Fairness Research}
\label{sec:future}
While there have been many definitions of, and approaches to, fairness in the literature, the study in this area is anything but complete. Fairness and algorithmic bias still holds a number of research opportunities. In this section, we provide pointers to outstanding challenges in fairness research, and an overview of opportunities for development of understudied problems.

\subsection{Challenges}
There are several remaining challenges to be addressed in the fairness literature. Among them are:
\begin{enumerate}
\item \textbf{Synthesizing a definition of fairness.} Several definitions of what would constitute fairness from a machine learning perspective have been proposed in the literature. These definitions cover a wide range of use cases, and as a result are somewhat disparate in their view of fairness. Because of this, it is nearly impossible to understand how one fairness solution would fare under a different definition of fairness. Synthesizing these definitions into one remains an open research problem since it can make evaluation of these systems more unified and comparable. having a more unified fairness definition and framework can also help with the incompatibility issue of some current fairness definitions.
\item \textbf{From Equality to Equity.} The definitions presented in the literature mostly focus on \emph{equality}, ensuring that each individual or group is given the same amount of resources, attention or outcome. However, little attention has been paid to \emph{equity}, which is the concept that each individual or group is given the resources they need to succeed~\cite{gooden2015race,mehrabi2020statistical}. Operationalizing this definition and studying how it augments or contradicts existing definitions of fairness remains an exciting future direction.
\item \textbf{Searching for Unfairness.} Given a definition of fairness, it should be possible to identify instances of this unfairness in a particular dataset. Inroads toward this problem have been made in the areas of data bias by detecting instances of Simpson's Paradox in arbitrary datasets~\cite{alipourfard2018wsdm}; however, unfairness may require more consideration due to the variety of definitions and the nuances in detecting each one. 
\end{enumerate}
\begin{figure}[h]
\includegraphics[width=\textwidth,trim=0cm 0cm 0cm 0cm,clip=true]{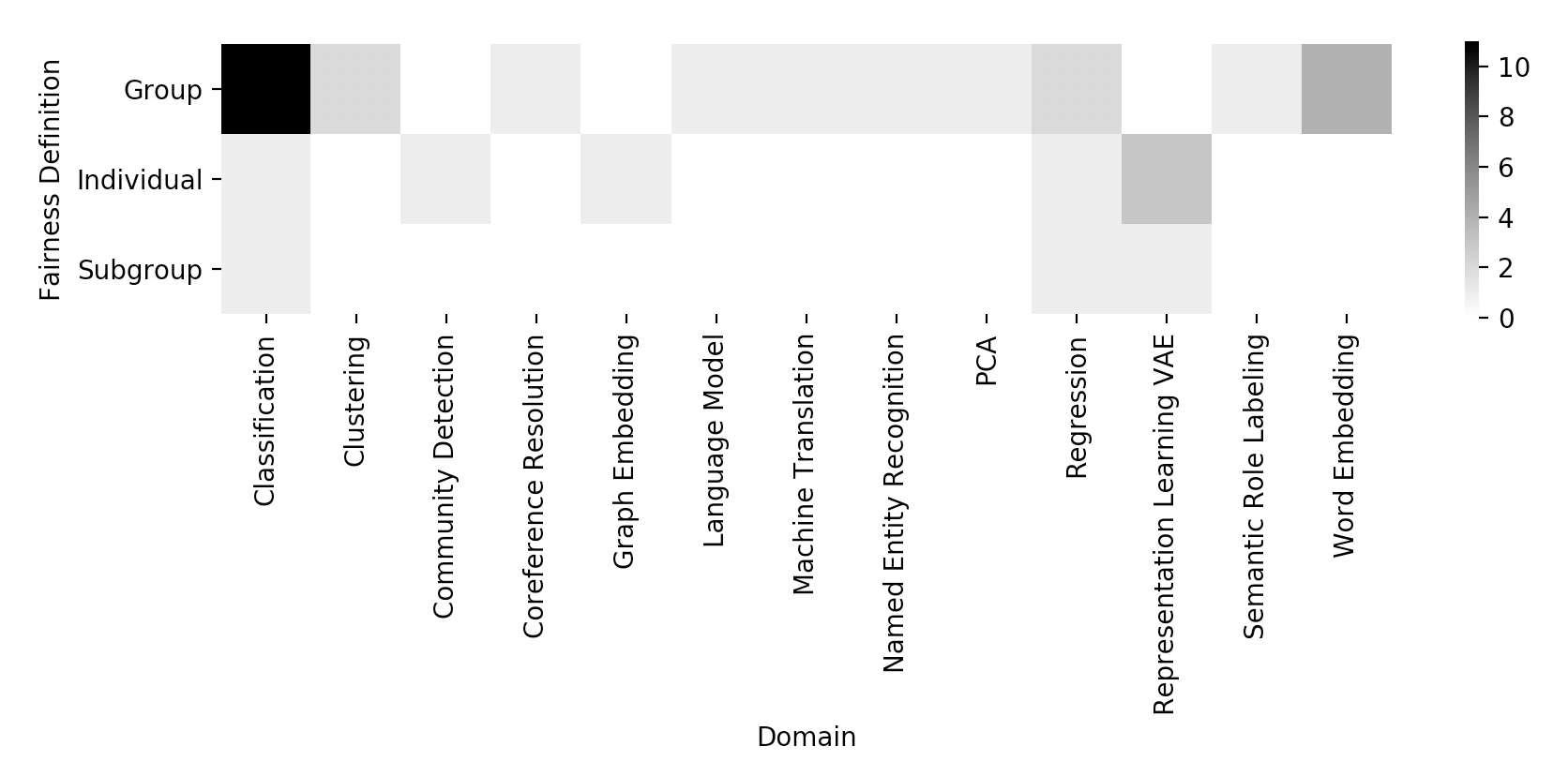}
\caption{Heatmap depicting distribution of previous work in fairness, grouped by domain and fairness definition.}
\label{heatmap}
\end{figure}
\subsection{Opportunities}
In this work we have taxonomized and summarized the current state of research into algorithmic biases and fairness---with a particular focus on machine learning. Even in this area alone, the research is broad. Subareas, from natural language processing, to representation learning, to community detection, have all seen efforts to make their methodologies more fair. Nevertheless, every area has not received the same amount of attention from the research community. Figure~\ref{heatmap} provides an overview of what has been done in different areas to address fairness---categorized by the fairness definition type and domain. Some areas (e.g., community detection at the subgroup level) have received no attention in the literature, and could be fertile future research areas.

\section{Conclusion}
In this survey we introduced problems that can adversely affect AI systems in terms of bias and unfairness. The issues were viewed primarily from two dimensions: data and algorithms. We illustrated problems that demonstrate why fairness is an important issue. We further showed examples of the potential real-world harm that unfairness can have on society---such as applications in judicial systems, face recognition, and  promoting algorithms. We then went over the definitions of fairness and bias that have been proposed by researchers. To further stimulate the interest of readers, we provided some of the work done in different areas in terms of addressing the biases that may affect AI systems and different methods and domains in AI, such as general machine learning, deep learning and natural language processing. We then further subdivided the fields into a more fine-grained analysis of each subdomain and the work being done to address fairness constraints in each. The hope is to expand the horizons of the readers to think deeply while working on a system or a method to ensure that it has a low likelihood of  causing potential harm or bias toward a particular group. With the expansion of AI use in our world, it is important that researchers take this issue seriously and expand their knowledge in this field. In this survey we categorized and created a taxonomy of what has been done so far to address different issues in different domains  regarding the fairness issue. Other possible future work and directions can be taken to address the existing problems and biases in AI that we discussed in the previous sections.
\section{Acknowledgments}
This material is based upon work supported by the Defense Advanced Research Projects Agency (DARPA) under Agreement No. HR0011890019. We would like to thank the organizers, speakers and the attendees at the IVADO-Mila 2019 Summer School on Bias and Discrimination in AI. We would like to also thank Brian Hu Zhang and Shreya Shankar.

\section{Appendix}
\subsection{Datasets for Fairness Research}
Aside from the existence of bias in datasets, there are datasets that are specifically used to address bias and fairness issues in machine learning. There are also some datasets that are introduced to target the issues and biases previously observed in older existing datasets. Below we list some of the widely known datasets that have the characteristics discussed in this survey.
\subsubsection{UCI Adult Dataset} UCI Adult dataset, also known as "Census Income" dataset, contains information, extracted from the 1994 census data about people with attributes such as age, occupation, education, race, sex, marital-status, native-country, hours-per-week etc., indicating whether the  income of a person exceeds \$50K/yr or not. It can be used in fairness-related studies that want to compare gender or race inequalities based on people's annual incomes, or various other studies \cite{Asuncion+Newman:2007}. 
\subsubsection{German Credit Dataset} The German Credit dataset contains 1000 credit records containing attributes such as personal status and sex, credit score, credit amount, housing status etc. It can be used in studies about gender inequalities on credit-related issues \cite{Dua:2019}. 
\subsubsection{WinoBias} The WinoBias dataset follows the winograd format and has 40 occupations in sentences that are referenced to human pronouns.
There are two types of challenge sentences in the dataset requiring linkage of gendered pronouns to either male or female stereotypical occupations. It was used in the coreference resolution study to certify if a system has gender bias or not---in this case, towards stereotypical occupations \cite{zhao2018gender}. 
\subsubsection{Communities and Crime Dataset} The Communities and Crime dataset gathers information from different communities in the United States related to several factors that can highly influence some common crimes such as robberies, murders or rapes. The data includes crime data obtained from the 1990 US LEMAS survey and the 1995 FBI Unified Crime Report. It also contains socio-economic data from the 1990 US Census.
\subsubsection{COMPAS Dataset} The COMPAS dataset contains records for defendants from Broward County indicating their jail and prison times, demographics, criminal histories, and COMPAS risk scores from 2013 to 2014 \cite{larson2016compas}. 
\subsubsection{ Recidivism in Juvenile Justice Dataset} The Recidivism in Juvenile Justice dataset contains all juvenile offenders between ages 12-17 who committed a crime between years 2002 and 2010 and completed a prison sentence in 2010 in Catalonia's juvenile justice system \cite{tolan2019machine}. 
\subsubsection{Pilot Parliaments Benchmark Dataset} The Pilot Parliaments Benchmark dataset, also known as PPB, contains images of 1270 individuals in the national parliaments from three European (Iceland, Finland, Sweden) and three African (Rwanda, Senegal, South Africa) countries. This benchmark was released to have more gender and race balance, diversity, and representativeness \cite{pmlr-v81-buolamwini18a}. 
\subsubsection{Diversity in Faces Dataset} The Diversity in Faces (DiF) is
an image dataset collected for fairness research in face recognition. DiF is a large dataset containing one million annotations for face images. It is also a diverse dataset with diverse facial features, such as different 
craniofacial distances, skin color, facial symmetry and contrast, age, pose, gender, resolution, along with diverse areas and ratios \cite{merler2019diversity}.
\begin{table}[H]
\centering
\begin{tabular}{ |p{5.3cm}||p{1.3cm}|p{3.3cm}|p{3.3cm}|}
 \hline
 Dataset Name& Reference & Size&Area\\
 \hline
UCI adult dataset& \cite{Asuncion+Newman:2007}&48,842 income records&Social\\[0.5pt]
 \hline
 German credit dataset&\cite{Dua:2019}&1,000 credit records&Financial\\[0.5pt]
 \hline
 Pilot parliaments benchmark dataset&\cite{pmlr-v81-buolamwini18a}&1,270 images &Facial images\\[0.5pt]
  \hline
 WinoBias&\cite{zhao2018gender}&3,160 sentences&Coreference resolution\\[0.5pt]
  \hline
 Communities and crime dataset&\cite{redmond2011communities}&1,994 crime records&Social\\[0.5pt]
  \hline
 COMPAS Dataset&\cite{larson2016compas}& 18,610 crime records&Social\\[0.5pt]
  \hline
  Recidivism in juvenile justice dataset&\cite{capdevila2005reincidencia}&4,753 crime records&Social\\[0.5pt]
  \hline
  Diversity in faces dataset&\cite{merler2019diversity}&1 million images&Facial images\\[0.5pt]
  \hline
\end{tabular}
\caption{Most widely used datasets in the fairness domain with additional information about each of the datasets including their size and area of concentration.}
\label{dataset}
\end{table}

\bibliographystyle{ACM-Reference-Format.bst}
\bibliography{references}
\end{document}